\documentclass{article}

\usepackage{arxiv}

\usepackage[utf8]{inputenc} 
\usepackage[T1]{fontenc}    
\usepackage{hyperref}       
\usepackage{url}            
\usepackage{booktabs}       
\usepackage{amsfonts}       
\usepackage{nicefrac}       
\usepackage{microtype}      
\usepackage{lipsum}		
\usepackage{graphicx}
\usepackage{natbib}
\usepackage{doi}

\usepackage{setspace}
\usepackage{graphicx} 
\usepackage{subcaption} 
\usepackage{amsmath,amsfonts,amsthm,bm}
\newtheorem{assumption}{Assumption} 
\usepackage{xcolor} 
\usepackage[toc,page]{appendix}
\usepackage{bbm}
\usepackage{txfonts}
\usepackage{booktabs} 
\usepackage{multirow} 
\usepackage{comment}
\usepackage{arydshln} 
\usepackage{natbib} 
\usepackage{adjustbox}
\usepackage{empheq} 
\usepackage{makecell}
\usepackage{tcolorbox} 
\usepackage{pifont} 

\theoremstyle{definition}
\newtheorem{definition}{Definition}

\usepackage{hyperref} 
\hypersetup{
    colorlinks=true,                
    breaklinks=true,                
    linkcolor= blue,                
    bookmarksopen=false,
    citecolor=blue,
    linkbordercolor=blue
}

\newcommand{\revised}[1]{\textcolor{black}{#1}}

\title{Uplift modeling with continuous treatments: \\A predict-then-optimize approach}

\author{ 
    Simon De Vos
\thanks{Corresponding author: \texttt{simon.devos@kuleuven.be}} 
\\
	KU Leuven\\
	\And
    Christopher Bockel-Rickermann\\
	KU Leuven\\
	\And
    Stefan Lessmann \\
	Humboldt University of Berlin \\
	\And
    Wouter Verbeke \\
	KU Leuven \\
}

\renewcommand{\shorttitle}{Uplift modeling with continuous treatments: A predict-then-optimize approach}

\hypersetup{
pdftitle={Uplift modeling with continuous treatments: A predict-then-optimize approach},
pdfsubject={q-bio.NC, q-bio.QM},
pdfauthor={David S.~Hippocampus, Elias D.~Striatum},
pdfkeywords={First keyword, Second keyword, More},
}

\begin{document}
\maketitle

\begin{abstract}
The goal of uplift modeling is to recommend actions that optimize specific outcomes by determining which entities should receive treatment. 
One common approach involves two steps:
    first, an inference step that estimates conditional average treatment effects (CATEs), and second, an optimization step that ranks entities based on their CATE values and assigns treatment to the top $k$ within a given budget.
    While uplift modeling typically focuses on binary treatments, many real-world applications are characterized by continuous-valued treatments, i.e., a treatment dose. 
This paper presents a predict-then-optimize framework to allow for continuous treatments in uplift modeling. 
    First, in the inference step, conditional average dose responses (CADRs) are estimated from data using causal machine learning techniques.
    Second, in the optimization step, we frame the assignment task of continuous treatments as a dose-allocation problem and solve it using integer linear programming (ILP). 
This approach allows decision-makers to efficiently and effectively allocate treatment doses while balancing resource availability, with the possibility of adding extra constraints like fairness considerations or adapting the objective function to take into account instance-dependent costs and benefits to maximize utility.
The experiments compare several CADR estimators and illustrate the trade-offs between policy value and fairness, as well as the impact of an adapted objective function.
This showcases the framework's advantages and flexibility across diverse applications in healthcare, lending, and human resource management. All code is available on \href{https://github.com/SimonDeVos/UMCT}{https://github.com/SimonDeVos/UMCT}.
\end{abstract}


\section{Introduction}

In many applications, decision-makers are interested in learning the causal effects of a treatment on a particular outcome \citep{holland1986statistics,pearl2009causal}.
A crucial aspect is response heterogeneity, where entities respond differently to treatments based on their characteristics \citep{rubin2005causal}.
In intervention-based decision-making, precisely this heterogeneity is valuable to leverage for finding a treatment assignment policy that optimizes a specific outcome variable, based on historical data, while adhering to certain constraints such as scarce resources.
Uplift modeling (UM) is a set of techniques to find such policies \citep{devriendt2018literature}. 
An often-used predict-then-optimize UM approach combines conditional average treatment effect (CATE) estimation with an optimization step, where CATE estimation can be considered an inference step as part of a larger decision-making process \citep{fernandez2022causal}.

Many real-world applications, as illustrated by Table~\ref{tab:intro_applications}, contain complexities beyond binary treatments and budget as the sole constraint and are better characterized by a treatment dose, i.e., where the intervention can be applied across a continuous range of values \citep{hirano2004propensity,holland2015optimal}. 
Moreover, continuous treatments have benefits for making treatment allocation possibly more effective and efficient compared to their binary counterpart because the marginal utility of treatment can vary with different dose levels, and the treatments can be allocated on fine granularity.

\begin{table}[tb]
    \centering
    \begin{adjustbox}{width=\textwidth}
    \begin{tabular}{ l|l l l l l }
    \toprule
    \textbf{\textit{Application}} & \textbf{Outcome}               & \textbf{Cont. Treatment }          & \textbf{Cost-sensitivity}                      & \textbf{Constraint}                \\
    \midrule
    \textit{Credit}      & Default rate          & Interest rate             & Loss given default \textit{(o)}       & Regul. compliance \textit{(f)}    \\ 
    \textit{Healthcare}  & Sickness              & Medication dose           & Medication price \textit{(t)}         & Equal access \textit{(f)}             \\ 
    \textit{HR}          & Employee retention    & Training hours            & Hourly opportunity cost \textit{(t)}  & Instructors \textit{(b)}              \\ 
    \textit{Maintenance} & Machine up-time       & Maintenance freq.         & Machine criticality \textit{(o)}      & Spare parts avail. \textit{(b)}       \\
    \bottomrule
    \end{tabular}
    \end{adjustbox}
\vspace{0.1cm}
    \caption{
        This table displays exemplary applications of a UM setting with continuous-valued treatments and their corresponding details. 
        \textit{Cost-sensitivity} can either refer to outcome benefits \textit{(o)} or treatment costs \textit{(t)}
        and \textit{Constraint} to budget \textit{(b)} or fairness \textit{(f)}.
        The examples provided are illustrative and not intended to be exhaustive or fully comprehensive.
    }
    \label{tab:intro_applications}
\vspace{-0.3cm}
    \rule{\textwidth}{.5pt}
\end{table}

While there is a growing body of literature focusing on conditional average dose response (CADR) estimation (i.e., the continuous-valued counterpart of CATE), 
this line of work focuses on causal inference and finding optimal doses,
largely ignoring the role of continuous treatments in constrained decision-making contexts \citep{bica2020estimating,schwab2020learning,nie2021vcnet}.
Constraints play a vital role in embedding business requirements into decision-making, which we accomplish using an integer linear programming (ILP) formulation.
While many requirements exist, fairness stands out as a key consideration and is the main exemplary constraint throughout this work \citep{barocas2023fairness}. 
Though fairness constraints are not new in machine learning for decision-making, they are often embedded as soft constraints during model training \citep{han2023retiring,frauen2023fair}. 
This approach, however, is limited in flexibility and modularity. 
By shifting the inclusion of fairness considerations to a post-processing step, i.e., after model training, decision-making is more flexible toward dynamic business requirements.

Therefore, in this paper, we develop a UM framework with continuous treatments.
Our work addresses two key gaps in the current literature. 
First, UM has not been formally defined, with the distinction between treatment effect estimation and allocation optimization largely overlooked. 
Second, as far as we are aware, there are no established methods in the literature extending UM to handle (i) continuous treatments,
let alone methods that manage the combined complexities of also including (ii) extra constraints (like fairness) and (iii) objective function alteration (to, for example, account for cost-sensitivity).

By addressing these gaps, our contributions are fourfold:

\begin{enumerate}
    \item 
    \revised{%
    We provide a framework that clearly defines UM, and how this differentiates from a mere causal inference problem, i.e., treatment effect estimation.
    }

    \item 
    \revised{%
    We extend UM methods to effectively handle continuous treatments by first leveraging state-of-the-art causal ML methods for CADR estimation and then defining the optimization part as a dose-allocation problem. The ILP formulation offers flexibility to adapt the objective function (e.g., to incorporate cost-sensitivity) and to include additional constraints (e.g., to enforce fairness considerations).
    }

    \item 
    \revised{%
    We are the first to include fairness considerations in a UM setting as explicit constraints in an ILP formulation. 
    By extension, to the best of our knowledge, this is also the first application of such constraints in decision-making pipelines that utilize ML predictions as input.
    }

    \item 
    \revised{%
    With a series of experiments, we show the capabilities of our framework and demonstrate how incorporating fairness constraints or cost-sensitive objectives influences policy outcomes, crucial for applications such as those highlighted in Table~\ref{tab:intro_applications}.
    }
\end{enumerate}

The remainder of this paper is structured as follows.
Section~\ref{sec:uplift_modeling} defines UM and elaborates on related literature.
Section~\ref{sec:problem} introduces the problem formulation and establishes the necessary notation.
Section~\ref{sec:methodology} focuses on the methodology, explaining the predict-then-optimize approach, the predictive models used to estimate CADRs, and the optimization techniques employed for constrained treatment allocation.
Section~\ref{sec:experiments} details the experimental setup, including data, evaluation metrics, and the experiments, followed by a discussion of the results.
Finally, Section~\ref{sec:conclusion} presents our conclusions, discusses limitations, and outlines potential further research.

\section{Uplift modeling}\label{sec:uplift_modeling}

In this section, we begin by discussing the purpose of UM, offering a clear definition and a detailed breakdown of its core elements. We then explore how UM can be extended to accommodate continuous treatments. Finally, we consider adjustments to UM, such as incorporating additional constraints or modifying the objective function, to better align with specific application requirements, reflecting its relevance to the domain of prescriptive analytics.

\subsection{Purpose and definition}

UM is a well-established set of methods in the field of personalized decision-making, closely aligned with the goals of prescriptive analytics \citep{devriendt2018literature}. 
Unlike traditional predictive models that identify entities likely to yield a desired outcome, UM distinguishes between baseline responders (entities likely to show positive outcomes even without intervention) and true responders that respond because of the treatment.
Consequently, UM focuses on outcome changes directly attributable to the treatment assignment, rather than simply predicting positive outcome probabilities. This helps avoid suboptimal targeting and ensures assignment policies capture their true incremental impact. 
Traditional UM predominantly deals with binary treatments, where the main goal is to rank individuals based on their expected response to a treatment \citep{zhang2021unified}. 
Works such as \citet{devriendt2018literature} provide a comprehensive overview, focusing on a setting with binary treatment effects, various modeling techniques, and the associated challenges. 
However, the literature often lacks a unified definition of UM, with some studies equating it to treatment effect inference \citep{gutierrez2017causal} while others focus primarily on ranking methods \citep{devriendt2020learning}.
To resolve this ambiguity, we propose the following UM definition:

\begin{definition}[]
\textbf{Uplift modeling} refers to the collection of methods where the \textcolor{black}{task} at hand is optimally allocating \textcolor{black}{treatments}, with the \textcolor{black}{objective} of maximizing the \textcolor{black}{total benefit} generated by these treatments, determined by the \textcolor{black}{cumulative uplift} over entities, under given  \textcolor{black}{constraints}.
\end{definition}

\revised{%
The above definition encompasses both the one-step approach (i.e., predict-and-optimize) and the two-step approach (i.e., predict-then-optimize). 
Notably, this definition is method‑agnostic to keep it inclusive of decision‑focused methods that learn policies directly and therefore do not output an explicit treatment‑effect estimate.
}
As Table~\ref{tab:uplift_positioning} highlights, the methodological contribution of this work focuses on the two-step approach, where treatment effect estimation is a component of the broader UM framework, which also includes treatment allocation as a critical task.

\begin{table}[tb]
    \centering
    \begin{tabular}{l | ll}
    \toprule
         \textit{\textbf{Treatment}}    & \textbf{Predict-and-optimize}     & \textbf{Predict-then-optimize}    \\
    \midrule
         \textit{Binary}       & Learn to rank    & \makecell[l]{\textit{(i)} CATE estimation \\ \textit{(ii)} Treatment-allocation problem}\\
    \cmidrule(lr){1-3}
         \textit{Continuous}   & Learn to allocate & \colorbox{lightgray}{\makecell[l]{\textit{(i)} CADR estimation \\ \textit{(ii)} Dose-allocation problem}} \\
    \bottomrule
    \end{tabular}
\vspace{0.1cm}
    \caption{A schematic overview of the positioning of our work within various UM approaches. 
    The focus of this paper is highlighted in grey.
    In the Decision-Focused Learning (DFL) paradigm, the goal is to directly include optimal treatment allocation in the model learning task (i.e., predict-and-optimize). 
    Prediction-Focused Learning (PFL) consists of (i) an inference and (ii) an optimization step (i.e., predict-then-optimize). 
    }
    \label{tab:uplift_positioning}
\vspace{-0.3cm}
    \rule{\textwidth}{.5pt}
\end{table}

\subsection{Treatment effects}

A treatment refers to an intervention or action that can be applied to an entity to influence a particular outcome. 
Treatments can be binary, where the action is \textit{treat} or \textit{do not treat} (e.g., offering or withholding discount). The CATE measures the expected difference in potential outcomes between treated and untreated groups, conditioned on entities' characteristics 
\citep{shalit2017estimating}. 
Additionally, recent developments have extended traditional UM to multi-treatment scenarios, which generalize the setup by considering more than two treatment options \citep{olaya2020survey} (e.g., offering discounts through different channels), 
and sequential treatments, where multiple treatments can be applied over time, each potentially influencing the outcome effects of subsequent treatments \citep{hatt2024sequential}.
However, these are not the focus of our paper and are considered future extensions.

The estimation of causal treatment effects from data is inherently challenging due to the fundamental problem of causal inference, namely, the absence of counterfactual outcomes \citep{holland1986statistics}. 
In the case of observational data, this problem is exacerbated by the possible non-random assignment of treatments and the resulting confounding bias \citep{pearl2009causal}. 
A real-life example is the overestimation of workplace wellness programs' impact on employee wellbeing, where non-random incentives cause treated populations to differ from the general population \citep{jones2019workplace}. A naive approach may overfit on the self-selected group, missing the true causal effect. This highlights how confounding can lead to issues like Simpson’s paradox, where aggregated data misrepresents trends compared to stratified data \citep{jones2019workplace}.
While randomized controlled trials (RCTs) are the gold standard for mitigating such biases \citep{verstraete2023estimating}, they are often impractical in real-world business settings because of cost, ethical, or operational reasons. 
Moreover, in the context of continuous treatments, with virtually infinite possible doses, setting up an RCT for dose-response estimation becomes significantly more complex \citep{holland2015optimal}.
In contrast, observational data is often cheap and readily accessible.
Therefore, methods that aim to balance the treatment and control groups have been proposed, including techniques like propensity score matching \citep{rosenbaum1983central} or covariate balancing \citep{johansson2016learning,shalit2017estimating}. 

In recent years, the consideration of continuous treatments—and consequently, CADR estimation—has gained traction because of its relevance in applications where understanding the effects of varying dose levels is critical (see Table~\ref{tab:intro_applications}).
Estimating CADRs is inherently more challenging than CATEs because it involves a spectrum of treatment doses, rather than values $0$ and $1$ \citep{schwab2020learning}. 
Also, with the theoretically infinite number of possible dose values, data sparsity and the non-uniformity of dose assignment could be challenging, making accurate effect estimation at each dose level difficult \citep{bockel2024sources}.
The Conditional Average Dose Effect (CADE), which can be directly derived from the CADR, generalizes the CATE for continuous treatments. 
Our paper specifically focuses on UM with this type of treatment.
Traditionally, dose effects have been derived from average dose-response curves using methods like the Hirano-Imbens estimator \citep{hirano2004propensity}, which extends propensity scores to continuous interventions through generalized propensity scores. However, these approaches often overlook individual-level heterogeneity in responses.
Recent advancements in ML have introduced techniques for learning individualized dose-response curves
\citep{bica2020estimating,schwab2020learning,nie2021vcnet,zhang2022exploring,bockel2023learning}. 
However, analogously to the case of CATE estimation, these methods primarily focus on the inference step and do not incorporate an optimization step where the learned dose-response relationships are used to prescribe decisions \citep{fernandez2022causal}.

\revised{
In this paper, we do not make any assumptions about the shape of the CADR. 
Instead, we adopt a flexible, data-driven approach.
In contrast, existing work, such as \citep{zhou2023direct},
relies on a strong structural assumption --- the law of diminishing marginal utility --- 
which implies that the marginal benefit of increasing the dose is strictly decreasing. 
While this assumption may hold in some applications, it limits model flexibility and generalizability. Our framework does not rely on such assumptions: the estimated dose-response relationships in our approach are non-parametric and can capture arbitrary, potentially non-monotonic shapes.
}

\subsection{Allocation task}

\subsubsection{Cumulative uplift \& total benefit as driver}

The allocation task aims to optimize an objective value driven by (cumulative) uplift. 
While uplift applies to a single entity, cumulative uplift reflects the policy's overall impact. 
For a policy treating $K$ entities, cumulative uplift is the sum of CATEs for discrete treatments or CADEs for continuous treatments across these treated entities.
In certain applications, cumulative uplift directly aligns with the target objective, as seen in contexts where maximizing cumulative uplift is the explicit goal \citep{devriendt2018literature}. 
However, cumulative uplift itself does not necessarily represent the objective. For example, in value-driven analytics, where uplift might disregard expected profit, and in cases where each entity may be associated with an individual benefit and treatment cost \citep{gubela2020response,lemmens2020managing,gubela2021uplift,verbeke2023or, haupt2022targeting}. 
Hence, when cumulative uplift is not the end goal in itself, allocations with lower cumulative uplift may yield a greater increase in objective value.

\subsubsection{Predict-and-optimize} 

In UM, the \textcolor{black}{task of allocating treatments} can follow one of two paradigms, represented by the two columns in Table~\ref{tab:uplift_positioning}.
On the one hand, there is the paradigm of decision-focused learning, where the optimal treatment allocation is directly learned in an integrated way \citep{betlei2021uplift, vanderschueren2024metalearners}. 
On the other hand, prediction-focused learning separates two distinct steps:
First, an inference task, where treatment effects are estimated, followed by an optimization task that uses these estimates as input to allocate treatments \citep{devriendt2021you}.

In this paper, we adopt the predict-then-optimize paradigm, which is represented in the second column of Table~\ref{tab:uplift_positioning}. With this approach, we clearly distinguish the treatment estimation task as a component of the larger UM approach, which additionally includes an optimization step that takes estimated treatment effects as input \citep{fernandez2022causal}.
In the case of binary treatments, the inference step involves estimating CATEs, followed by an optimization step, i.e., finding a policy by solving the treatment-allocation problem to determine which entities should receive treatment.
Typically, this is done using a greedy ranking heuristic: entities are ranked based on their predicted CATEs, and treatments are assigned to the top-$k$ entities until the budget is exhausted \citep{jaskowski2012uplift}. Alternatively, under a flexible budget, the optimal value of $k$ is determined through cost-benefit analysis \citep{verbeke2023or}.

In contrast, the predict-and-optimize approach, or decision-focused learning (DFL), integrates prediction and optimization into a single end-to-end system \citep{elmachtoub2022smart}. 
This paradigm tailors predictions specifically for downstream objectives, using regret-based loss functions to align predictions with optimization goals, where \(\mathbf{x}^{\star}(\mathbf{c})\) represents the optimal decision with full information, and \(\mathbf{x}^{\star}(\hat{\mathbf{c}})\) is based on estimated parameters \citep{mandi2023decision}:
\begin{equation}
    \mathcal{L}_{regret} = f\left(\mathbf{x}^{\star}(\mathbf{c}), \mathbf{c}\right) - f\left(\mathbf{x}^{\star}(\hat{\mathbf{c}}), \mathbf{c}\right)
    \label{eq:L_regret}
\end{equation}
%

Although DFL approaches for UM exist \citep{betlei2021uplift, vanderschueren2024metalearners}, we adopt the predict-then-optimize framework due to its simplicity, flexibility, stability, and generalizability to other treatment types. 
\revised{
It allows using various predictive models without being restricted by a specific optimization setup, which is especially useful for complex or pre-trained models without access to the training process \citep{barocas2023fairness}. 
Moreover, it enables flexible adjustments to the objective function and integration of additional constraints other than budget limits, such as fairness considerations, allowing for different allocations under varying budget conditions.
Our approach requires modifying only the optimization step, leaving the prediction model unchanged and does not require a complete model retraining.
}

\subsubsection{Allocation problem as ILP}

When considering continuous treatments, the optimization step becomes a \textit{dose-allocation problem}, where the task is to determine the optimal treatment dose for each entity. 
This generalizes the binary \textit{treatment-allocation problem}, which can often be addressed by ranking heuristics. 
However, since for the \textit{dose-allocation problem} the decision space expands, a greedy ranking heuristic is no longer feasible, and a more formal optimization approach is required.

Therefore, we formulate and solve the allocation problem as an ILP.
To consider binary choices per dose level, we discretize the estimated CADRs into distinct dose options.
This discretization is motivated by two factors: (i) it allows the ILP formulation and (ii) many practical applications only permit pre-defined dose levels with limited granularity \citep{zhan2024weighted}.
While this discretization introduces some loss of generalization, its impact is minimal since the number of dose bins is flexible and only affects the optimization step without altering the prediction step. 
The number of dose bins can be freely adjusted, although more bins increase computational complexity.

\subsubsection{Flexible optimization constraints and objectives}

During the optimization step, constraints and objective functions can be flexibly adjusted. Any constraint compatible with ILP can be added, and the objective function can be adapted, as long as it remains a linear combination of decision variables, as required by the ILP. Fairness constraints and cost-sensitive objective functions are exemplary, and further modifications, such as operational constraints, can also be incorporated

\paragraph{Fairness as constraint}

The importance of fairness in algorithmic decision-making is well-recognized, especially as AI systems are increasingly deployed in high-stakes domains like criminal justice, hiring, and lending \citep{chouldechova2020snapshot,kozodoi2022fairness,barocas2023fairness}. 
Algorithmic bias can lead to unfair outcomes, as demonstrated in notable cases like the COMPAS tool for recidivism prediction \citep{dressel2018accuracy} and issues with Amazon's hiring algorithms, which were scrapped \citep{dastin2022amazon}. 
Group fairness concepts include independence, ensuring predictions are unaffected by sensitive attributes; separation, requiring conditional independence of predictions given the outcome; and sufficiency, mandating conditional independence of the outcome given the prediction \citep{makhlouf2021applicability, barocas2023fairness}.
The EU’s AI Act underscores the regulatory focus on fairness, aiming to prevent AI from reinforcing discrimination through a risk-based framework \citep{european_ai_act_2021}. 
Our framework’s flexibility in setting fairness constraints enables its use across various risk categories, ensuring compliance with these requirements and balancing fairness with utility \citep{european_ai_act_2021}.

Balancing utility --- such as profit and accuracy --- against fairness, like equal treatment across demographics, remains a key challenge. \citet{kleinberg2016inherent} and \citet{corbett2017algorithmic} show that enhancing fairness often trades off with utility and that multiple fairness definitions may even conflict.
Traditional fairness assessments emphasize output-based metrics, such as demographic parity or equal opportunity, which measure group-level outcome disparities \citep{barocas2023fairness}.
Recent approaches consider the entire distribution of predictions or decisions, incorporating statistical moments and exploring distributional fairness \citep{khan2023fairness,han2023retiring}.
In decision-making, fairness extends beyond prediction to both allocation fairness—ensuring treatment levels are independent of sensitive attributes like race or gender—and outcome fairness, which ensures equitable results across groups. 
\revised{
Although outcome predictions may be imperfect, enforcing fairness at the level of expected outcomes has been shown to be an effective and ethically justified approach in decision-making settings \citep{scantamburlo2024prediction,corbett2017algorithmic}.
While not ideal, these estimates still represent the best available proxy for assessing how different groups will benefit from the allocation policy.
}
Fairness notions, which may conflict, can be incorporated as hard constraints during optimization \citep{corbett2017algorithmic}, similar to this work's approach, or as soft constraints in multi-objective learning \citep{barocas2023fairness}.
Recent research, including \citet{frauen2023fair}, has largely focused on fairness in observational data with binary sensitive attributes, while \citet{nabi2019learning} apply causal inference to multi-stage decision-making under fairness constraints.
However, these studies do not address fairness in the context of continuous treatments, a critical gap our research aims to fill.

\paragraph{Value-driven objective function}

In managerial decision-making, profit maximization or cost reduction is often the primary goal. Cost-sensitive or value-driven methods are crucial as they incorporate asymmetric costs and benefits, aligning decisions with these business objectives \citep{hoppner2022instance}. 
Cost-sensitive approaches seek to balance costs and benefits, an aspect often ignored by standard methods \citep{gubela2020response,verbeke2023or}. 
In this work, costs and benefits are deterministic while assuming stochastic treatment effects, setting it apart from studies where both costs and benefits are modeled as stochastic \citep{verbraken2012novel,verbeke2012new}.
Cost-sensitive methods find applications across various domains, including credit risk \citep{bahnsen2014example}, fraud detection \citep{vasquez2022hidden,de2023robust}, customer churn \citep{verbraken2012novel}, business failure prediction \citep{de2020cost}, and machine maintenance \citep{vanderschueren2023optimizing}.

\section{Problem formulation}\label{sec:problem}

\subsection{Notation}

\begin{figure}[t]
    \centering
    \includegraphics[width=0.20\linewidth]{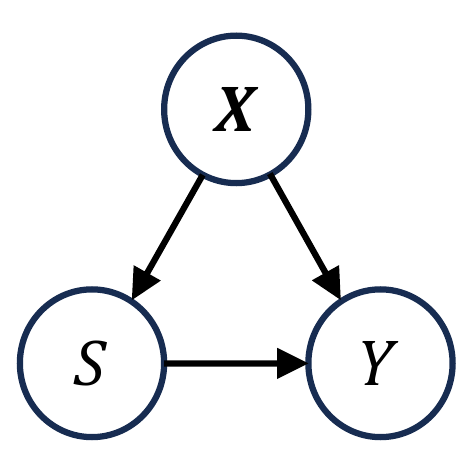}\\
    \caption{This DAG represents the assumed causal relationships between variables in the training data. $\boldsymbol{X}$: entity's pre-treatment features, $S$: treatment dose, $Y$: outcome.}
    \rule{\textwidth}{.5pt}
    \label{fig:dag}
\end{figure}

The dataset \(\mathcal{D} = \{(\mathbf{x}_i, s_i, y_i)\}_{i=1}^{N}\) has \(N\) tuples of pre-treatment features $\mathbf{X} \in \mathcal{X} \subset \mathbb{R}^k$, treatment doses $S \in \mathcal{S} = \left[0,1\right]$, and outcomes $Y \in \mathcal{Y} =\left[0,1\right]$, where \(\mathbf{x}_i\), \(s_i\), and \(y_i\) denote the respective values for instance \(i\).
We adopt the Rubin-Neyman potential outcomes framework \citep{rubin2004direct, rubin2005causal}, originally proposed for binary treatments $Y(0)$ and $Y(1)$, which can be extended to multi-valued or continuous treatments $Y(S)$ across \(S \in [0,1]\) \citep{schwab2020learning}. For each instance, the potential outcome \(y_i(s)\) reflects the response to dose \(s\), given features \(\mathbf{x}_i\).       
Data tuples \((\mathbf{x}, s, y)\) are generated by distributions \(p(\mathbf{X})\), \(p(S)\), and the observed policy \(\Pi_{obs}\), which assigns treatment \(s_i\) based on \(\mathbf{x}_i\), potentially introducing confounding bias. Figure \ref{fig:dag} illustrates the assumed causal structure.
The CADR function \(\mu: \mathcal{S} \times \mathcal{X} \rightarrow [0,1]\) is defined as \(\mu(s, \mathbf{x}) = \mathbb{E}[Y(s) \mid \mathbf{X} = \mathbf{x}]\), representing the expected outcome for a given dose \(s\) and features \(\mathbf{x}\). 
The CADE function \(\tau: \mathcal{S} \times \mathcal{X} \rightarrow [-1,1]\) is then derived from the CADR and defined as \(\tau_s(\mathbf{x}) = \mathbb{E}[Y(s) - Y(0) \mid \mathbf{X} = \mathbf{x}]\), which measures the difference in expected outcomes between dose \(s\) and the baseline dose \(0\).
Our notation is summarized in \ref{app:notation}. 

\subsection{Prediction step}

During the prediction step, our goal is to train a model that can accurately estimate $\hat{y}(s)$, providing unbiased estimates of $\mu(s, \mathbf{x})$ and $\tau_s(\mathbf{x})$ across the entire domain of $S \in [0,1]$. 
The estimated CADR for an instance with features $\mathbf{x}$ is defined as $\hat{\mu}(s, \mathbf{x})=\mathbb{E}\left[\hat{Y}(s) \mid \mathbf{X}=\mathbf{x}\right]$. The estimated CADE for a given dose $s$ is defined as $\hat{\tau}_s(\mathbf{x}) = \hat{\mu}(s, \mathbf{x})-\hat{\mu}(0, \mathbf{x}) = \mathbb{E}\left[\hat{Y}(s)-\hat{Y}(0) \mid \mathbf{X}=\mathbf{x}\right]$.
To use the CADE estimates as input for the optimization step, we discretize the CADRs into $\delta$ bins.
For a given $\delta$, we define $D = \left\{\left.\frac{d-1}{\delta} \right\rvert\, d=1, \ldots, (\delta+1)\right\}$ so that the vector $\hat{\bm{\tau}}(\mathbf{x}) = \Bigl(\hat{\tau}_{D_d}(\mathbf{x})\Bigr)_{d=1}^{\delta}$ then contains an entity's $\delta$ CADE estimates.
\revised{We discuss model training and outcome estimation in Section~\ref{sec:methodology:cadr_estimation}.}

\subsection{Optimization step}

Let $\pi: \mathcal{X} \rightarrow \{0,1\}^{\delta}$ be an 
assignment policy defined for a single entity that takes its features as input and assigns a dose to this entity. In this policy output, the $d^{th}$ element equals $1$ if dose $s$ is assigned, and $0$ otherwise. Dose $s$ corresponds to element $d$ in $D$ (i.e., $D_d = s$). 

The cost matrix \(\mathbf{C} \in \mathbb{R}^{N \times \delta}\) defines treatment costs, where \(c_{i,d}\) is the treatment cost for entity \(i\) at dose \(s\), with \(c_{i,0} = 0\). 
\revised{For each entity \(i,\) its row in the cost matrix is denoted by $\mathbf{C}_i \;=\; (c_{i,0},\, c_{i,1},\, \dots,\, c_{i,\delta - 1})$, which lists all possible costs for that entity.}
\revised{For simplicity, we assume costs are directly proportional to dose levels (e.g., a dose of $0.2$ has a cost of $0.2$). This cost function is modular and can be easily replaced with alternative cost structures to suit different scenarios or requirements.}
The cost of policy $\pi$ for instance \(i\) is:

\begin{align}
    \Psi_i(\pi)=\left\langle\pi\left(\mathbf{X}_i\right), 
    \mathbf{C}_{i\cdot}
    \right\rangle
\end{align}

\revised{
The uplift for an instance \(i\) ($U_i$) is given by the dot product $\langle\cdot,\cdot\rangle$ between the assignment policy $\pi$ and the CADE vector $\bm{\tau}$.
Similarly, the value gain for instance \(i\) ($V_i$) is the same dot product, scaled by the instance-specific benefit \(b_i\).
}

\revised{
Due to the fundamental problem of causal inference, historical data provides only factual outcomes, not counterfactual outcomes. Thus, in realistic scenarios, we must estimate these counterfactual outcomes. In contrast, experimental settings can leverage semi-synthetic datasets where the ground truth is known because we control the data-generating process. For a detailed discussion on the fundamental problem of causal inference, we refer readers to \cite{pearl2009causal}.
We define three distinct assignment policy values, each depending on the availability and usage of ground-truth versus estimated CADR values.
}

\revised{
The expected policy value  (Eq.~\ref{eq:Vi_exp}) uses estimated CADRs for both the assignment policy and evaluation. It represents the anticipated value achievable under realistic conditions without access to counterfactual ground-truth data.
The prescribed policy value \(V_i^{presc}\) (Eq.~\ref{eq:Vi_presc}) employs estimated CADRs to determine the assignment policy but uses ground-truth CADRs for evaluation. Policies determined by maximizing \(V_i^{exp}\) yield an observable policy value \(V_i^{presc}\). Due to differences between estimated and true causal effects, \(V_i^{presc}\) typically diverges from \(V_i^{exp}\).
The optimal policy value \(V_i^{opt}\) (Eq.~\ref{eq:Vi_opt}) serves as a benchmark and uses ground-truth CADRs for both policy determination and evaluation. This represents the maximum attainable policy value in a hypothetical setting where all counterfactual dose-response information is known precisely, a condition unattainable in practical applications.
}

Formally, these are defined as:

\begin{align}
\revised{
U_i^{exp}
} 
&
\revised{
= \langle \pi(\hat{\tau}(\mathbf{X}_i)), \hat{\tau}(\mathbf{X}_i) \rangle, \label{eq:Ui_exp} 
}
\\
    \revised{
    U_i^{presc} 
    }
    &
    \revised{
    = \langle \pi(\hat{\tau}(\mathbf{X}_i)), \tau(\mathbf{X}_i) \rangle, 
    }
    \label{eq:Ui_presc} 
    \\
    \revised{
    U_i^{opt} 
    }
    &
    \revised{
    = \langle \pi(\tau(\mathbf{X}_i)), \tau(\mathbf{X}_i) \rangle.
    }\label{eq:Ui_opt}
\end{align}

\begin{align}
    V_i^{exp}   &= U_i^{exp} b_i 
                = \langle \pi(\hat{\tau}(\mathbf{X}_i)), \hat{\tau}(\mathbf{X}_i) \rangle b_i, \label{eq:Vi_exp} \\
    V_i^{presc} &= U_i^{presc} b_i 
                = \langle \pi(\hat{\tau}(\mathbf{X}_i)), \tau(\mathbf{X}_i) \rangle b_i, \label{eq:Vi_presc} \\
    V_i^{opt}   &= U_i^{opt} b_i 
                = \langle \pi(\tau(\mathbf{X}_i)), \tau(\mathbf{X}_i) \rangle b_i. \label{eq:Vi_opt}
\end{align}


For $N$ entities with features $\mathbf{X}$ and a budget $B$, the policy $\stackrel{B}\Pi: \mathbb{R^+}\times\mathcal{X}^N \rightarrow \{0,1\}^{(N\times\delta)}$ can be based on estimated or ground-truth CADRs. 
The expected optimal policy $\stackrel{B}\Pi^{exp}$ (Eq.~\ref{eq:Pi_exp}) maximizes value within the budget using estimated CADRs. 
The prescribed policy assigns treatments according to $\stackrel{B}\Pi^{exp}$, but uses ground-truth CADRs. 
The ground-truth optimal policy $\stackrel{B}\Pi^{opt}$ (Eq.~\ref{eq:Pi_opt}), based solely on ground-truth CADRs, is the full-information benchmark.

\begin{align}
    \stackrel{B}\Pi^{exp} = \operatorname{argmax}\left\{\sum_{i=1}^N V_i^{exp}(\pi): \sum_{i=1}^N \Psi_i(\pi) \leq B\right\}\label{eq:Pi_exp} \\
    \stackrel{B}\Pi^{opt} = \operatorname{argmax}\left\{\sum_{i=1}^N V_i^{opt}(\pi): \sum_{i=1}^N \Psi_i(\pi) \leq B\right\}\label{eq:Pi_opt}
\end{align}

\revised{The implementation of this optimization formulation is discussed in Section~\ref{sec:methodology:ilp}.}

\section{Methodology}\label{sec:methodology}

\begin{figure}[t]
\centering
    \begin{subfigure}[b]{0.32\textwidth}
    \includegraphics[width=\textwidth]{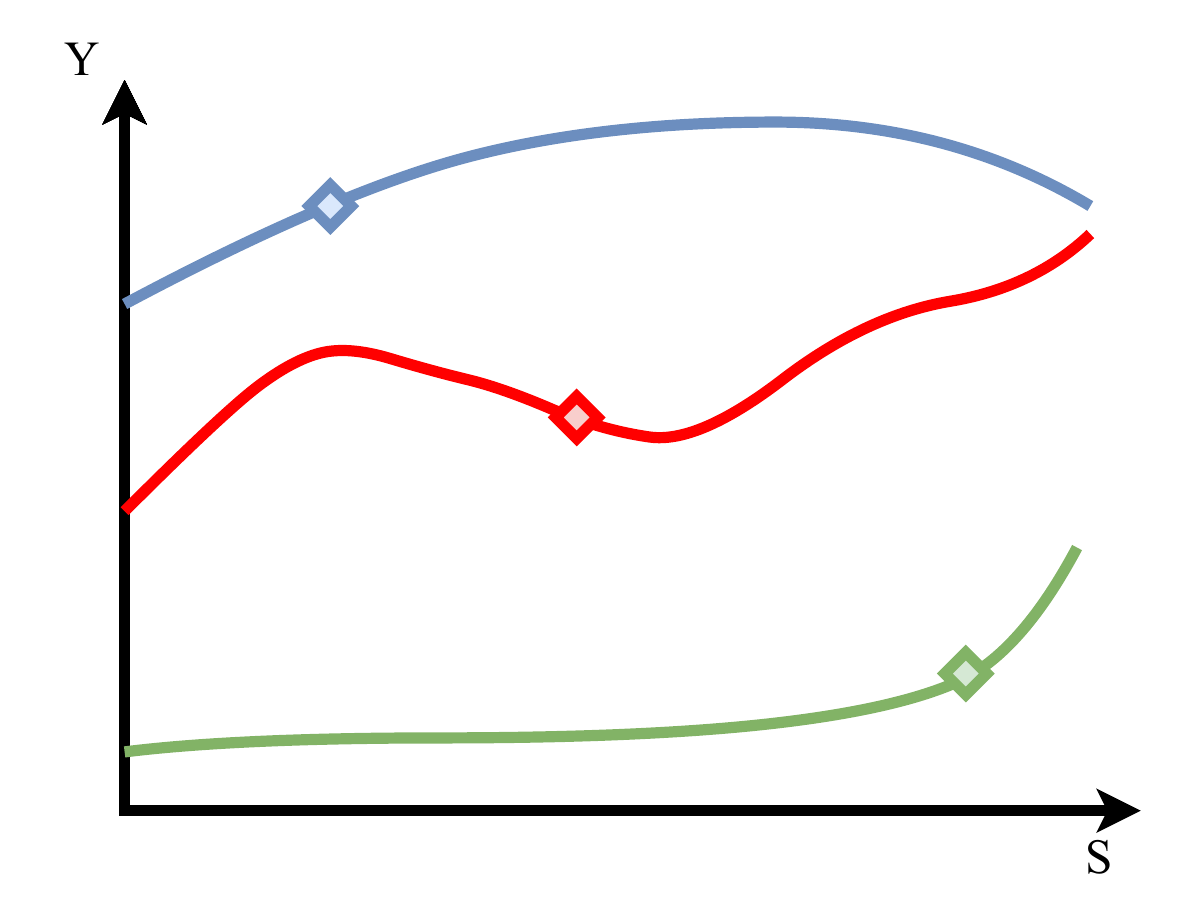}
    \caption{Data: Factual observations sampled from ground-truth dose-response curves}
    \label{fig:method_overview1}
\end{subfigure}
\hfill
\begin{subfigure}[b]{0.32\textwidth}
    \includegraphics[width=\textwidth]{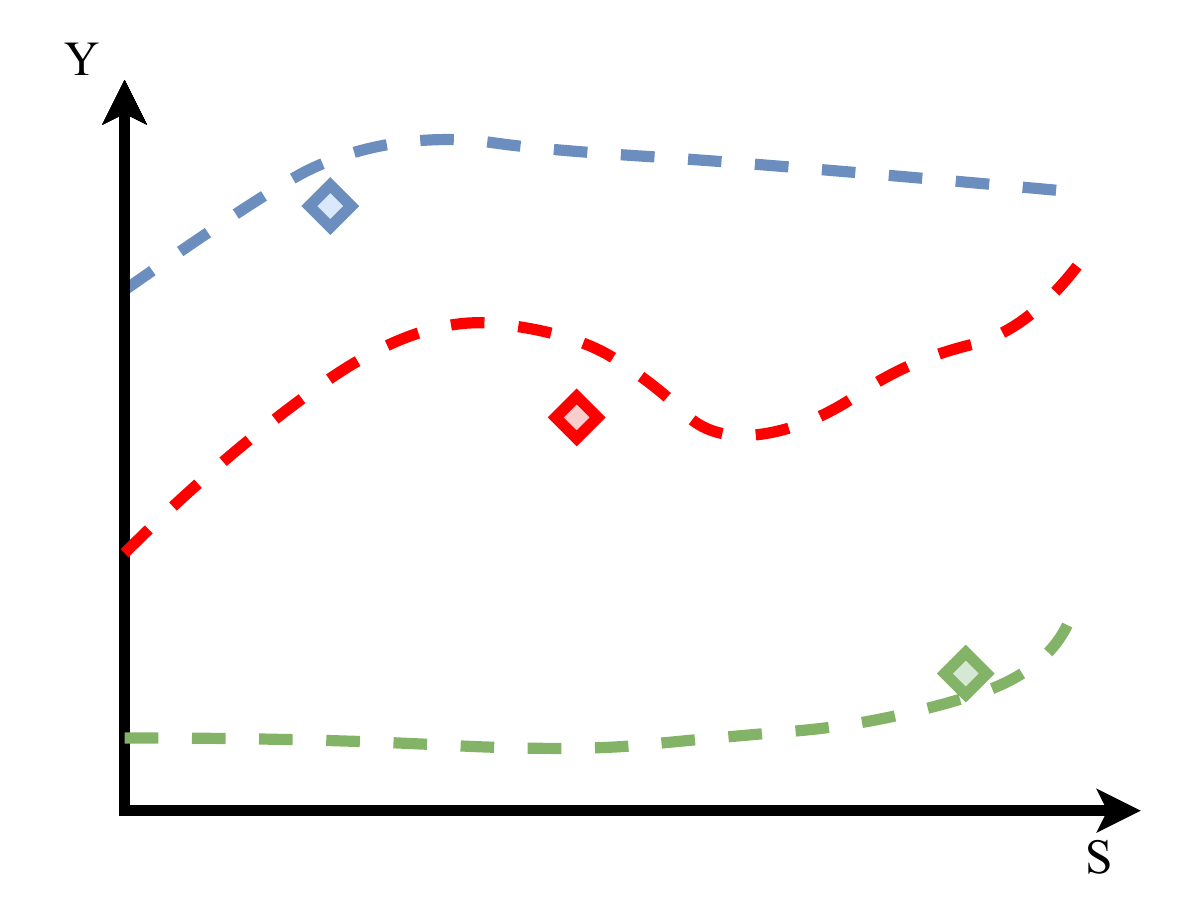}
    \caption{Prediction step: Dose-response estimation based on factual observations}
    \label{fig:method_overview2}
\end{subfigure}
    \hfill
\begin{subfigure}[b]{0.32\textwidth}
    \includegraphics[width=\textwidth]{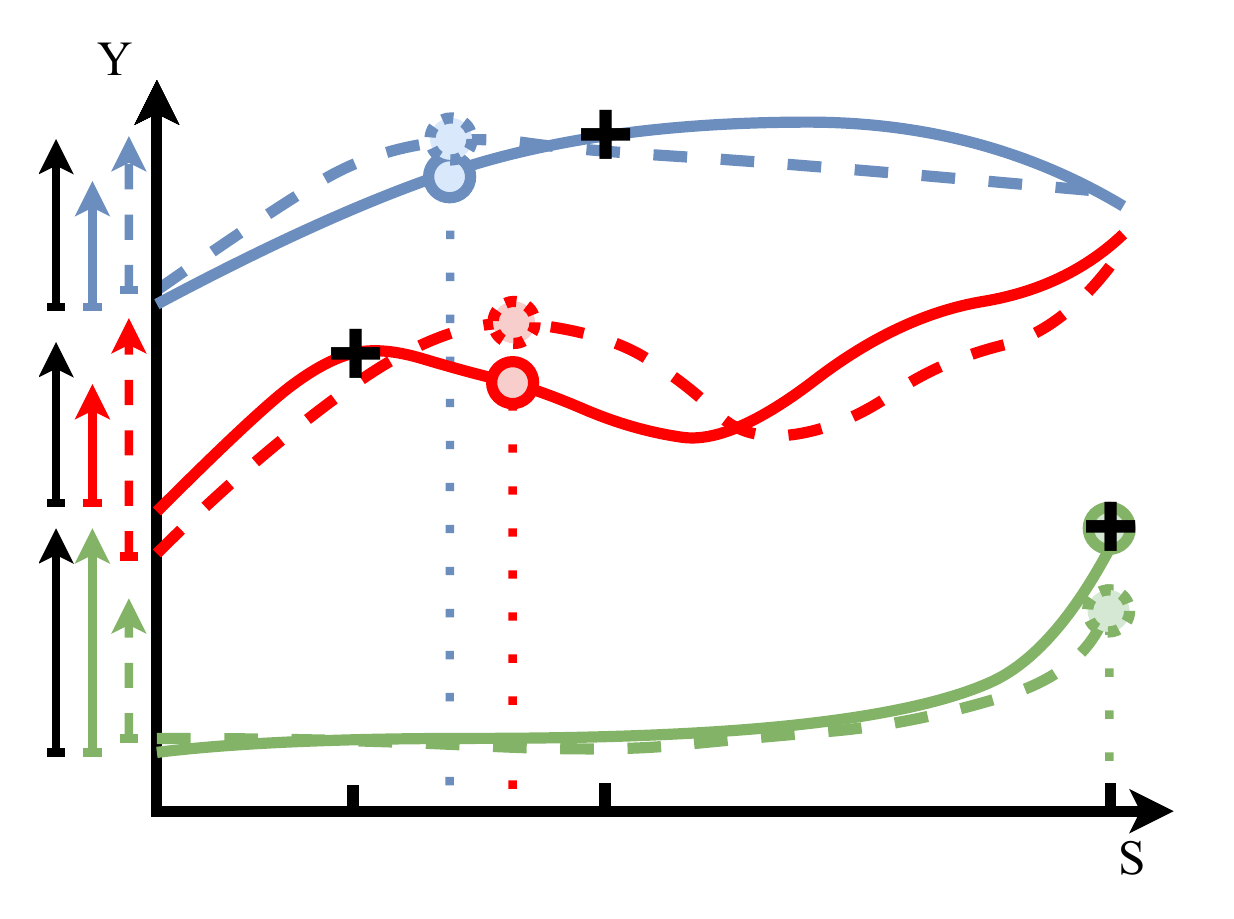}
    \caption{Optimization step: Treatment dose allocation based on estimated dose-responses}
    \label{fig:method_overview3}
\end{subfigure}

\caption{Overview of predict-then-optimize approach for UM with continuous treatment effects for three entities. 
The x-axis represents the dose $S$, the y-axis represents the outcome $Y$. Key elements include true dose-response curves (solid lines), estimated dose-response curves (dashed lines), prescribed policies with their estimated and true outcomes (resp. dotted and full-lined circles), and the corresponding expected and prescribed values (resp. dotted and full arrows in color). The black \ding{58}'s and arrows correspond to the full-information solution.}
\label{fig:method_overview}
\rule{\textwidth}{.5pt}
\end{figure}

Figure~\ref{fig:method_overview} summarizes the predict-then-optimize approach for UM with continuous treatments under constraints.

\subsection{Predictive model for CADR estimation}\label{sec:methodology:cadr_estimation}

To accurately estimate CADRs $\hat{\mu}$ and their associated CADE estimations $\hat{\tau}$, not addressing confounding between $S$ and $\boldsymbol{X}$ in observational data can lead to detrimental inaccuracies or biases \citep{pearl2009causal}. 
A variety of methods for disentangling treatment effects from confounding factors have been proposed in the literature \citep{johansson2016learning, schwab2020learning, bockel2023learning}, and this is not a prime concern in this paper. 
Instead, we consider the predictive methods as off-the-shelf solutions, with methods that do not apply a debiasing scheme also being applied in the experiments.
One may refer to, e.g., \citet{bica2020estimating}, for assessments of the impact of confounding on outcomes.

\revised{
Following standard practices in causal inference, we make three assumptions for estimating potential outcomes from observational data: Consistency, Ignorability, and Overlap \citep{imbens2000role, lechner2001identification}. 
These assumptions ensure the identifiability of the CADE function $\tau_s(\mathbf{x})$, i.e., they result in Equation~\ref{eq:identified} (see proof in~\ref{app:assumptions}).
}

\begin{assumption}\label{ass:consistency}
    \textit{Consistency:} $\forall s \in \mathcal{S}: Y = Y(s)$. 
    This means that for any entity with observed treatment dose $S = s$, the potential outcome for this treatment dose is equal to the factual observed outcome.
\end{assumption}
\begin{assumption}\label{ass:ignorability}
    \textit{Ignorability:} $ \left\{ Y(s) \mid s \in \mathcal{S}\right\} \Perp S \mid \mathbf{X}$. 
    This means that, conditional on an entity's pre-treatment characteristics, the assigned treatment dose $S$ is independent of the potential outcomes.
\end{assumption}
\begin{assumption}\label{ass:overlap}
    \textit{Overlap:} $\forall \mathbf{x} \in \mathcal{X} \text{ \textit{such that} } p(\mathbf{x})>0, \forall s \in \mathcal{S}: 0 < p(s \mid \mathbf{x}) < 1 $. 
    {This states that all entities had a non-zero probability of being assigned any dose.}
\end{assumption}

In this work, we compare four different learning methods to estimate $\hat{\mu}$: 
an S-learner with random forests as the base learner (S-Learner (rf)), 
an S-learner with a feedforward multi-layer perceptron (MLP) without any debiasing as the base learner (S-Learner (mlp)) \citep{kunzel2019metalearners}, 
DRNet \citep{schwab2020learning}, 
and VCNet \citep{nie2021vcnet}. 
Further details on the hyperparameters can be found in \ref{app:hyperparameters}.

\subsection{ILP for the dose-allocation problem}\label{sec:methodology:ilp}

Deciding on the optimal treatment dose for each entity, represented by policies $\stackrel{B}\Pi^{exp}$ and $\stackrel{B}\Pi^{opt}$, is equivalent to solving Equations~(\ref{eq:Pi_exp}-\ref{eq:Pi_opt}).  
%
In the case when only binary doses can be applied, i.e., $S \in \{0,1\}$, this problem is an instance of the knapsack problem \citep{dantzig1957discrete}.
A sensible heuristic would be to rank allocations from high to low uplift until the budget capacity is reached. This approach mirrors the calculation of a traditional Qini curve in a binary treatment setting.
%
%
\revised{
In our more general setting, however, any dose \(S \in [0,1]\) is permitted. To formalize this, we introduce two related functions:
\[
\stackrel{}U: \{0,1\}^{(N\times\delta)} \times [-1,1]^{(N\times\delta)} \rightarrow \mathbb{R}
\quad\text{and}\quad
\stackrel{}V: \{0,1\}^{(N\times\delta)} \times [-1,1]^{(N\times\delta)} \times \mathbb{R}^N \rightarrow \mathbb{R}.
\]
Both take as input a policy \(\stackrel{B}\Pi \in \{0,1\}^{(N\times\delta)}\) (expressed as a matrix of decision variables) and a matrix \(\mathbf{T} \in [-1,1]^{(N\times\delta)}\) of CADE values (estimated \(\hat{\tau}\) or ground-truth \(\tau\)) for the \(N\) entities. The function \(\stackrel{}V\) additionally requires a benefit vector \(\mathbf{b} \in \mathbb{R}^N\).
Note that $\stackrel{}U$ is a special case of $\stackrel{}V$ in which the benefit vector $\mathbf{b}$ is set to all ones. For simplicity, we elaborate on $\stackrel{}V$, with the understanding that this case also encompasses $\stackrel{}U$.
}

%
\revised{
To solve Equations~(\ref{eq:Pi_exp}--\ref{eq:Pi_opt}), we propose the following ILP (Equations \ref{eq:ILP_formulation_begin}-\ref{eq:ILP_formulation_end}), where \(V_i(\pi)\) in the objective is replaced by \(V_i^{exp}\) when finding \(\stackrel{B}\Pi^{exp}\) and by \(V_i^{opt}\) when finding \(\stackrel{B}\Pi^{opt}\).
Additionally, we compare the ILP to a greedy ranking heuristic (see Section~\ref{sec:exp1})
}
Although numerous business requirements exist, we focus on fairness as the main example of constraints throughout this work.
Without loss of generalization, we consider one protected binary sensitive attribute $A$ 
\revised{(which is also included as an input feature to train the predictive model)}, 
where $N_{\mathrm{A=0}}$ is the index set of entities where the protected attribute $A=0$, and $N_{\mathrm{A=1}}$ represents those with $A=1$. These two sets are mutually exclusive and collectively exhaustive, so that $|N_{\mathrm{A=0}}| + |N_{\mathrm{A=1}}| = |N|$.
To address fairness in allocation and outcomes, we introduce two slack parameters: $\epsilon_{DT} \in [0,1]$ for allocation fairness and $\epsilon_{DO} \in [0,1]$ for outcome fairness. A slack parameter value of $\epsilon=0$ guarantees perfect fairness between the groups $N_{\mathrm{A=0}}$ and $N_{\mathrm{A=1}}$, while parameter $\epsilon=1$ removes this fairness constraint.

\begin{alignat}{2}\label{eq:ILP_formulation_begin}
    \text{max} \quad 
    & \sum_{i=1}^N V_i(\pi) & \\
    \text{s.t.} \quad 
    & \sum_{i=1}^N \Psi_i(\pi) \leq B 
    & \quad & \text{(Budget constraint)} \\
    & \frac{1}{|N_{\mathrm{A=0}}|} \sum_{i \in N_{\mathrm{A=0}}} \sum_{s \in D} \pi_s(X_i) \cdot s 
    \ge (1 - \epsilon_{DT}) \cdot 
    \frac{1}{|N_{\mathrm{A=1}}|} \sum_{i \in N_{\mathrm{A=1}}} \sum_{s \in D} \pi_s(X_i) \cdot s 
    & \quad & \text{(Alloc. fairness 1)} \\
    & \frac{1}{|N_{\mathrm{A=0}}|} \sum_{i \in N_{\mathrm{A=0}}} \sum_{s \in D} \pi_s(X_i) \cdot s 
    \le (1 + \epsilon_{DT}) \cdot 
    \frac{1}{|N_{\mathrm{A=1}}|} \sum_{i \in N_{\mathrm{A=1}}} \sum_{s \in D} \pi_s(X_i) \cdot s 
    & \quad & \text{(Alloc. fairness 2)} \\
    & \frac{1}{|N_{\mathrm{A=0}}|} \sum_{i \in N_{\mathrm{A=0}}} \sum_{s \in D} \pi_s(X_i) \cdot \tau_s 
    \ge (1 - \epsilon_{DO}) \cdot 
    \frac{1}{|N_{\mathrm{A=1}}|} \sum_{i \in N_{\mathrm{A=1}}} \sum_{s \in D} \pi_s(X_i) \cdot \tau_s 
    & \quad & \text{(Outc. fairness 1)} \\
    & \frac{1}{|N_{\mathrm{A=0}}|} \sum_{i \in N_{\mathrm{A=0}}} \sum_{s \in D} \pi_s(X_i) \cdot \tau_s 
    \le (1 + \epsilon_{DO}) \cdot 
    \frac{1}{|N_{\mathrm{A=1}}|} \sum_{i \in N_{\mathrm{A=1}}} \sum_{s \in D} \pi_s(X_i) \cdot \tau_s 
    & \quad & \text{(Outc. fairness 2)} \\
    & \sum_{s \in D} \pi_s(X_i) = 1, \forall i \in \{1, \ldots, N\} 
    & \quad & \text{(Exactly one dose)} \\
    & \pi_s(X_i) \in \{0,1\}, \forall s \in D, \forall i \in \{1, \ldots, N\}  
    & \quad & \text{(Binary decisions)} \label{eq:ILP_formulation_end}
\end{alignat}

Analogous to $V_i(\pi)^{exp}$, $V_i(\pi)^{presc}$, $V_i(\pi)^{opt}$ (Equations~(\ref{eq:Vi_exp}-\ref{eq:Vi_opt}), which are defined on instance-level), and Equations~(\ref{eq:Pi_exp}-\ref{eq:Pi_opt}) in Section \ref{sec:problem}, we consider three versions of $\stackrel{}V$ each differing in their dependence on ground-truth or estimated CADEs:

\begin{align}
\revised{
    \stackrel{}V^{exp} = \sum_{i=1}^N \sum_{d=1}^\delta \bigl(\stackrel{B}{\Pi}^{exp}_{i,d} \cdot \hat{T}_{i,d} \cdot b_i \bigr),
    }
    \label{eq:V_exp} \\
\revised{    
    \stackrel{}V^{presc} = \sum_{i=1}^N \sum_{d=1}^\delta \bigl(\stackrel{B}{\Pi}^{exp}_{i,d} \cdot {T}_{i,d} \cdot b_i \bigr),    
    }
    \label{eq:V_presc} \\
\revised{    
    \stackrel{}V^{opt} = \sum_{i=1}^N \sum_{d=1}^\delta \bigl(\stackrel{B}{\Pi}^{opt}_{i,d} \cdot {T}_{i,d} \cdot b_i \bigr).    
    }
    \label{eq:V_opt}
\end{align}

\section{Experiments}\label{sec:experiments}

In this section, we demonstrate our framework using semi-synthetic data, aiming to illustrate its applicability and performance in a controlled setting.
The main objective is to illustrate the working of both the prediction step, i.e., the inference of continuous treatment effects, and the optimization step.
We introduce relevant performance metrics and evaluate the impact of adding fairness constraints and adjusting the objective function on policy value.

\subsection{Data}\label{sec:data}

Our experiments use a semi-synthetic approach based on a real dataset about infant health and development programs (IHDP) \citep{brooks1992effects}.
This dataset originates from a randomized controlled trial and is frequently used as a benchmark for binary CATE estimation methods \citep{shalit2017estimating, guo2020survey}. 
Since we focus on continuous treatments, we adopt a semi-synthetic approach where both doses and outcomes are artificially generated. 
For this purpose, we follow the established literature \citep{nie2021vcnet}. 
\revised{
Detailed information about the original dataset and the semi-synthetic data generation process is provided in \ref{app:ihdp}.
}

\subsection{Evaluation metrics}

The predict-then-optimize approach is evaluated in two distinct steps, with separate metrics used to assess each step.

\paragraph{Prediction step}
To measure the accuracy of the predictive model in estimating individual dose-response curves, we use the \textit{Mean Integrated Squared Error} (MISE, Equation~\ref{eq:mise}) \citep{silva2016observational,schwab2020learning}. 
This metric evaluates how closely the predicted dose-response curves match the true dose-response functions over the entire range $\mathcal{S}$ and hence requires information on semi-synthetic ground-truth.

\begin{equation}\label{eq:mise}
    \operatorname{MISE}=\frac{1}{N} \sum_{i=1}^N \int_{s \in \mathcal{S}}\bigl(\mu\left(s, \mathbf{x}_i\right)-\hat{\mu}\left(s, \mathbf{x}_i\right)\bigr)^2 ds
\end{equation}

\paragraph{Optimization step}
In the optimization step, several metrics evaluate how well the treatments are allocated, using the estimated CADEs, under the imposed constraints.
\revised{
The primary metric is the total Value ($V$), computed as the sum of individual $V_i$ across all $N$ entities. We distinguish $V^{exp}$ (Eq. \ref{eq:V_exp}), $V^{presc}$ (Eq. \ref{eq:V_presc}), and $V^{opt}$ (Eq. \ref{eq:V_opt}) corresponding to the expected, prescribed, and optimal values, respectively.
}
The cost-insensitive version, where costs and benefits are equal over all entities, is noted as $\stackrel{}U$.
\textit{Regret} is defined as the difference between the full-information optimal value and the value achieved by the prescribed decision \citep{mandi2023decision}. 
In other words, it measures how suboptimal the prescribed decision is compared to the optimal solution under ground-truth parameters.
When CADRs are perfectly estimated, regret is zero. 
However, even with imperfect CADR estimations, regret can still be zero if the assignment policy remains optimal despite inaccuracies in the CADR estimates.
\begin{equation}
    Regret = V^{opt} - V^{presc} = V\left(\stackrel{B}{\Pi} (\tau), \tau\right) - V\left(\Pi (\hat{\tau}), \tau\right)
\end{equation}
To make $Regret$ scale-independent, we also include the normalized version:
\begin{equation}
    RegretN_{B} = \frac{\stackrel{}V^{opt} - \stackrel{}V^{presc}}{\stackrel{}V^{opt}} = \frac{V\left(\stackrel{B}\Pi (\tau), \tau\right) - V\left(\stackrel{B}\Pi (\hat{\tau}), \tau\right)}{V\left(\stackrel{B}\Pi (\tau), \tau\right)}
\end{equation}
Additionally, we assess performance across a range of budgets.
The \textit{Value Curve} plots the function $V(B) = \sum_{i=1}^N V_i(\pi_{B})$, for $B \in [0, B_{max}]$, where $B_{max}$ is defined as the budget required when treating all entities with dose $s=1$.
We report the \textit{Area Under the Value Curve} (AUVC) when costs and benefits are instance-dependent and \textit{Area Under the Uplift Curve} (AUUC) when costs and benefits are equal for all instances.
In a traditional binary treatment setting, the horizontal axis of the uplift curve typically represents the `cumulative proportion of entities targeted' \citep{jaskowski2012uplift}. In our case, this translates to `budget used' and differs in that the budget can target entities with finer granularity. For example, while a value of 5 on the x-axis in the traditional setting corresponds to 5 entities targeted, in our case, a budget of 5 could represent 10 entities receiving smaller doses.

\subsection{Results and discussion}

In this section, we describe the experimental evaluation of our proposed predict-then-optimize framework, focusing on its ability to handle continuous treatments,  optimize dose allocation under various constraints, and balance possibly conflicting objectives such as fairness and policy value.
In all experiments, the number of bins $\delta$ is fixed to 10, motivated by \ref{app:delta}, and the level of confounding bias is constant.
All experiments are implemented using \textit{Python 3.9} and can be reproduced with the code available on \href{https://github.com/SimonDeVos/UMCT}{Github}. All ILPs are solved using \textit{Gurobi 11.0}.
\revised{
In all experimental scenarios presented, the ILP solver converged, guaranteeing optimal solutions for the given problem instances. 
Solver convergence was systematically verified by checking the solver status after each optimization run.
}

\subsubsection{Experiment 1: Performance of dose-response estimators}\label{sec:exp1}

The first experiment aims to compare the performance of different dose-response estimators in terms of both prediction accuracy and capacity for good dose allocation.
We evaluate four dose-response estimators:
S-Learner (rf) and
S-Learner (mlp) \citep{kunzel2019metalearners}, 
DRNet \citep{schwab2020learning}, 
and VCNet \citep{nie2021vcnet}. 
For each estimator, we conduct an internal 5-fold cross-validation loop, tuning hyperparameters across a grid search (details in \ref{app:hyperparameters}). The model with the lowest average MSE on factual outcomes over all folds is selected.
In this experiment, we do not consider cost-sensitivity or fairness constraints. 
The budget is incrementally raised to observe the corresponding variation in $U^{presc}_B$. 
Each model's estimated CADRs are visualized in \ref{app:exp1_cadr}.

\begin{figure}[t]
    \centering
    \includegraphics[width=0.75\linewidth]{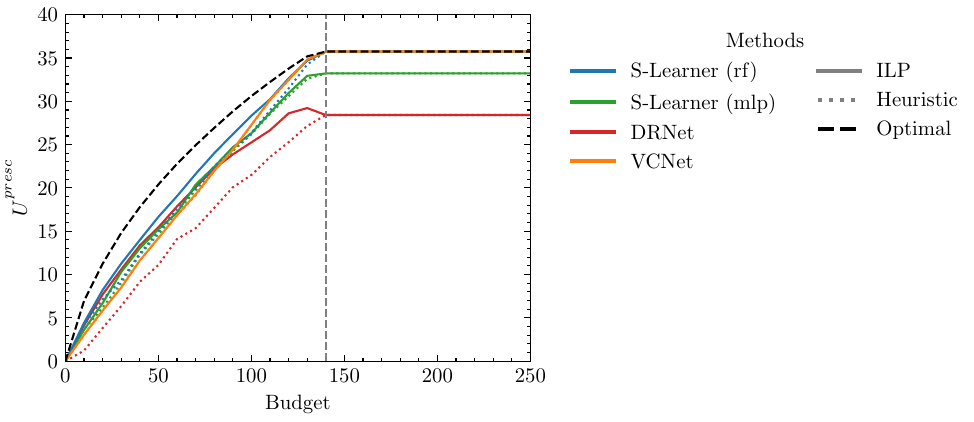}\\
    \caption{
        This figure shows $U^{presc}$ for four dose-response estimators as the available budget levels increase. \revised{Doses are allocated with an ILP and a heuristic approach.}
    }
    \label{fig:exp1_qini}
    \rule{\textwidth}{.5pt}
\end{figure}

\revised{
We allocate treatments using two distinct methods. 
The first approach utilizes the ILP introduced in Section~\ref{sec:methodology:ilp}. 
The second employs a heuristic inspired by the multi-treatment framework \citep{olaya2020survey}. Here, each discrete dose is treated as a separate intervention and, formally, the optimal treatment for an entity with features $\mathbf{x}_i$ is $\operatorname{argmax}(\mathbf{\hat{\tau}}(\mathbf{x}_i))$.
Entities are then ranked according to the uplift of their optimal treatment, and doses are assigned greedily in that order until the budget is depleted.   
}
The results are displayed in Figure~\ref{fig:exp1_qini} and Table~\ref{tab:exp_1}. The analysis provides two insights.

\begin{table}[t]
    \centering
    \begin{tabular}{l|ccccc}
        \toprule

            \multicolumn{1}{l}{\multirow{2}{*}{\textit{\textbf{Estimator}}}} & \multirow{2}{*}{\textbf{MISE}} & \multicolumn{2}{c}{\multirow{1}{*}{\textbf{AUUC@140}}} & \multicolumn{2}{c}{\multirow{1}{*}{\textbf{AUUC@250}}} \\

                                                            &                   &   \textit{Heur}. & \textit{ILP} & \textit{Heur}. & \textit{ILP} \\

            \midrule     
        
            \textit{S-Learner (rf)}  & 0.060            & \textbf{0.837}    & \textbf{0.892}    & \textbf{0.926}    & \textbf{0.951}    \\
            \textit{S-Learner (mlp)} & \textbf{0.044}   & 0.813             & \textit{0.829}    & 0.877             & 0.884             \\
            \textit{DRNet}           & \textit{0.049}   & 0.649             & 0.799             & 0.729             & 0.797             \\
            \textit{VCNet}           & 0.055            & \textit{0.827}    & 0.827             & \textit{0.922}    & \textit{0.922}    \\
        
\cdashline{1-6} 
           
            \textit{Optimal}             & 0.000             & 1.000             & 1.000  & 1.000             & 1.000                 \\
        \bottomrule
    \end{tabular}
\vspace{0.1cm}
     \caption{This table shows the MISE (prediction step) and AUUC for budgets 140 and 250 (optimization step) across four dose-response estimators. 
    The AUUC values are normalized against the optimal full-information solution.
    The best results are highlighted in bold and the second-best in italics.}
    \label{tab:exp_1}
\vspace{-0.3cm}
    \rule{\textwidth}{.5pt}
\end{table}

\revised{
First, compared to typical uplift curves from binary treatment scenarios (based on a ranking heuristic), both the multi-treatment heuristic and the ILP with continuous treatments as input do not fully utilize the budget.
Around a budget of 140, they find a point where further treatment becomes counterproductive ---i.e., it is more valuable to avoid \textit{overtreating} certain entities--- due to the non-monotonic nature of CADRs.
From a budget of 140 onwards, per estimator, the heuristic and ILP approach find the same solution (i.e., selection of the dose with maximum estimated CADE) and they converge in terms of $U^{presc}$.
}
Unlike the heuristic approach, which ranks binary treatment effects from high to low and selects the top $k$ within the budget, continuous treatment effects have a larger search space, allowing for partial treatments.        
This is evident in Figure~\ref{fig:exp1_qini} where the Uplift curves flatten around a budget of 140.
Therefore, in Table~\ref{tab:exp_1}, we include AUUC not only for the full budget (i.e., @250) but also for a partial budget of 140. 
Around this point, for all methods considered (and the full-information optimal solution), the policy value remains constant despite increasing budget.

Second, Table~\ref{tab:exp_1} shows a misalignment between the quality of CADR estimation (measured by MISE during prediction) and the quality of treatment allocation (measured by AUUC during the optimization step). 
MISE does not fully capture downstream task performance, as errors in critical areas of the curve have a greater impact on treatment allocation than those in less important regions 
\revised{--- which holds true for both heuristic and the ILP approaches}. 

\revised{
For example, although S-Learner (mlp) has the lowest MISE, its final allocations rank second or third. DRNet shows the second-best MISE but performs worst in terms of AUUC at both budget points, irrespective of the treatment assignment method. Conversely, S-Learner (rf), despite its high MISE, outperforms all others in uplift effectiveness.
}
This suggests that the best inference methods do not always result in optimal decision-making, a notion similarly observed in cost-sensitive learning literature, where maximizing predictive accuracy does not necessarily yield the highest profit \citep{verbeke2012new}.
Figure~\ref{fig:app_exp1_cadr_slearner} illustrates this phenomenon: the S-Learner (rf) produces low-quality estimates for $S\in[0.4,0.6]$, resulting in a relatively high MISE. However, errors in this region have minimal impact on optimization since they correspond to low CADEs and are not selected anyway. 
\revised{
DRNet, on the other hand, has relatively small errors overall but struggles significantly in the crucial high-impact region of $S\in[0.9,1.0]$, where the ground-truth CADR sharply declines (see Figure~\ref{fig:app_exp1_cadr_drnet}). 
Thus, accurate dose allocation doesn't strictly require perfect predictions, provided the predictions capture the critical areas effectively.
}
Conversely, perfect treatment allocation does not require flawless predictions; effective allocation is possible even with imperfect dose-response estimates.
\revised{
A further analysis of runtime for increasing problem sizes is provided in \ref{app:exp1_scalability}. While the heuristic approach scales well for increasing problem sizes, it does not allow additional side constraints like fairness, which are allowed for the ILP.
}

\subsubsection{Experiment 2: Fairness trade-offs in treatment allocation}\label{sec:exp2}

\begin{figure}[t]
    \centering
        \begin{subfigure}[b]{0.35\textwidth}
        \includegraphics[width=\textwidth]{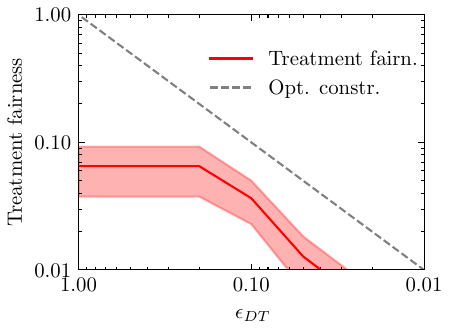}
        \caption{Allocation fairness}
        \label{fig:exp2_1}
    \end{subfigure}
    \begin{subfigure}[b]{0.35\textwidth}
        \includegraphics[width=\textwidth]{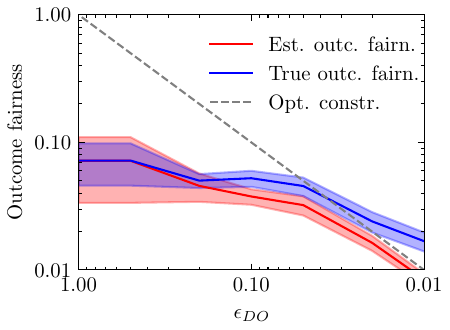}
        \caption{Outcome fairness}
        \label{fig:exp2_2}
    \end{subfigure}\\
    \caption{The effect of slack parameters $\epsilon_{DT}$ and $\epsilon_{DO}$ on, respectively, disparate treatment (allocation fairness) and disparate outcome (outcome fairness).
    Both panels use a logarithmic scale.
    }\label{fig:exp2_12}
    \rule{\textwidth}{.5pt}
\end{figure}

\begin{figure}[t]
    \centering
    \includegraphics[width=0.35\linewidth]{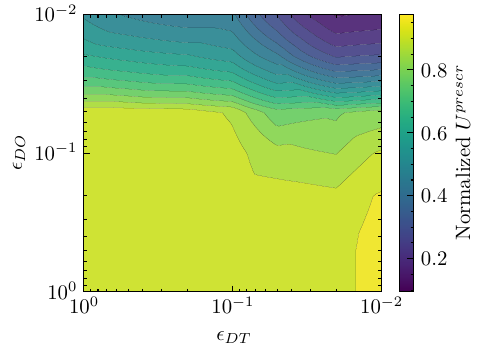}\\
    \caption{
    The x-axis represents the fairness constraint parameter for disparate treatment ($\epsilon_{DT}$), and the y-axis represents the fairness constraint parameter for disparate outcome ($\epsilon_{DO}$). Lower values of $\epsilon$ indicate stricter fairness constraints. 
    $U^{presc}$ is normalized between 0 and 1 and is aggregated over multiple budgets (from 25 to 250 in increments of 25). The averaged normalized $U^{presc}$ is color-coded, where blue means lower and yellow means higher.}
    \rule{\textwidth}{.5pt}
    \label{fig:exp2_3}
\end{figure}

\begin{figure}[t]
    \centering
    \begin{subfigure}{0.32\textwidth}
        \includegraphics[width=\textwidth]{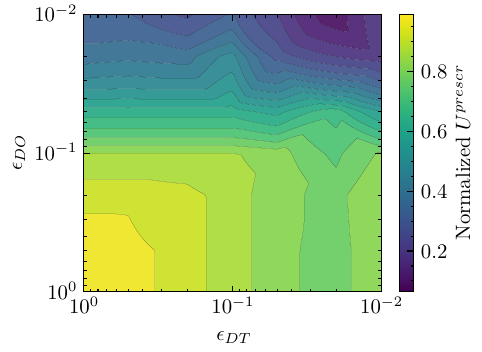}
        \caption{$\gamma=1$}
        \label{fig:exp2_4_01}
    \end{subfigure}   
    \hfill
    \begin{subfigure}{0.32\textwidth}
        \includegraphics[width=\textwidth]{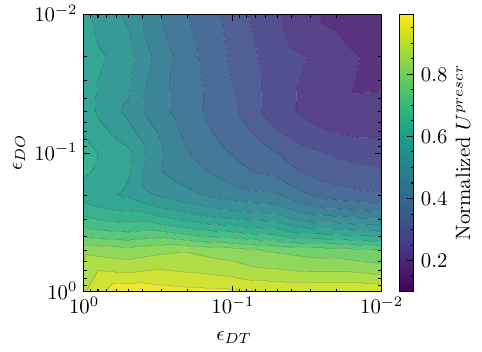}
        \caption{$\gamma=5$}
        \label{fig:exp2_4_05}
    \end{subfigure}
    \hfill
    \begin{subfigure}{0.32\textwidth}
        \includegraphics[width=\textwidth]{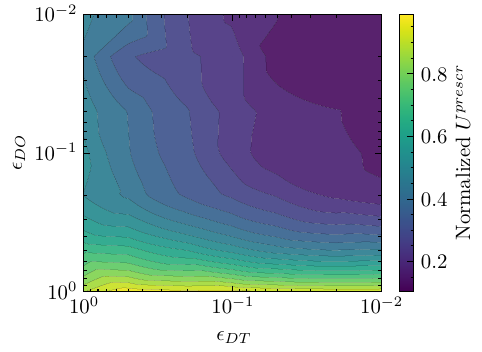}
        \caption{$\gamma=10$}
        \label{fig:exp2_4_10}
    \end{subfigure}         
    \caption{Fairness constraints and varying ground-truth ATEs.
    This figure examines the link between the effect of the protected feature on the ground-truth ATE and the \textit{ease} of achieving fairness. 
    One panel is provided for each value of $\gamma \in \{1, 5, 10 \}$ (the case for $\gamma=0$ is displayed in Figure~\ref{fig:exp2_3}) and is aggregated over multiple budgets (from 25 to 250 in increments of 25), using the S-learner with random forests as the base learner. The averaged normalized $U^{presc}$ is color-coded where blue means lower and yellow means higher.}
    \rule{\textwidth}{.5pt}
    \label{fig:exp2_4}
\end{figure}

The second experiment examines the trade-offs between policy value and business requirements as constraints, with a particular focus on fairness.
Specifically, we examine both allocation fairness and outcome fairness, analyzing the impact of tightening or loosening these fairness constraints on overall policy value.
In this experiment, we fix the budget and disregard cost-sensitivity. We use the S-Learner (rf), which performed best in terms of AUUC, as the dose-response estimator. 
We vary two key fairness constraints: allocation fairness, which measures the disparity in assigned doses between groups, and outcome fairness, which reflects the difference in treatment outcomes. 
These constraints are regulated by two slack parameters: $\epsilon_{DT}$ for allocation fairness and $\epsilon_{DO}$ for outcome fairness.

Figure~\ref{fig:exp2_1} shows the effect of parameter $\epsilon_{DT}$ on treatment fairness. 
Since treatment allocation is deterministic, there is no difference between the ground truth and estimates. 
Treatment fairness can be adjusted to any desired level by controlling $\epsilon_{DT}$, making it straightforward to manage within the given constraints.

\revised{%
Figure~\ref{fig:exp2_2} examines outcome fairness as regulated by the parameter $\epsilon_{DO}$, revealing a distinct challenge compared to treatment fairness: a discrepancy between estimated and ground truth fairness. This divergence arises because optimization is based on estimated CADEs rather than true values. Although tightening $\epsilon_{DO}$ (illustrated by the red line) can restrict estimated disparities, it does not guarantee alignment with ground truth fairness (blue line). This misalignment is driven by the quality of the CADE estimates and underscores a broader challenge in predictive analytics—namely, that fairness in predictions does not always translate into fairness in actual outcomes. Nevertheless, we argue that incorporating constraints on expected outcomes across groups remains a principled and ethically sound design choice. Even when individual-level outcomes are uncertain, they still provide the most informative basis available for assessing how different groups may benefit from the allocation policy.
}

Figure~\ref{fig:exp2_3} presents the trade-off between fairness constraints and $U^{presc}$. 
Stricter enforcement of both treatment and outcome fairness leads to lower uplift. 
The effects of these constraints differ; without an outcome constraint, the treatment constraint can be tightened with barely harming the policy value. 
Conversely, tightening the outcome constraint reduces policy value, especially when the treatment constraint is also strict.
Note that finding a solution to the ILP to satisfy both fairness constraints is always possible, with the trivial available option of not allocating any treatment.

The setup of Figure~\ref{fig:exp2_4} is similar to the one in Figure~\ref{fig:exp2_3}, but also examines varying data-generating processes where the two groups defined by the protected feature are more different. This is controlled by the parameter $\gamma$, with higher values indicating greater differences in ground-truth average treatment effects (ATEs) between the two protected groups.
Figure~\ref{fig:exp2_4} plots the results for three increasing values of $\gamma$ (the case for $\gamma=0$ is displayed in Figure~\ref{fig:exp2_3}). As the difference in ATEs between the groups increases, the effect of slack parameters $\epsilon_{DT}$ and $\epsilon_{DO}$ on the objective value (normalized $U^{presc}$, averaged over different budgets) becomes more pronounced.
Strict slack parameters are tolerable without significant value loss when the groups are similar ($\gamma=0$, Figure~\ref{fig:exp2_3}). In contrast, even mild slack parameters significantly affect the objective value when ATEs are most different ($\gamma=10$, Panel \ref{fig:exp2_4_10}).

\subsubsection{Experiment 3: The impact of cost-sensitivity on utility}\label{sec:exp3}

\revised{
In many applications, the ultimate goal is profit or cost reduction, not just uplift. This experiment demonstrates that optimizing for the true objective (value) yields different policies than optimizing for uplift alone.
}
This experiment explores the effect of incorporating cost-sensitivity into policy optimization. Building upon the previous experiments, we now account for cost-sensitivity in the adapted objective function of ILP, considering instance-dependent treatment costs, $\mathbf{C}$, and outcome benefits, $\mathbf{b}$.
\begin{figure}[t]
    \centering
        \begin{subfigure}[b]{0.35\textwidth}
        \includegraphics[width=\textwidth]{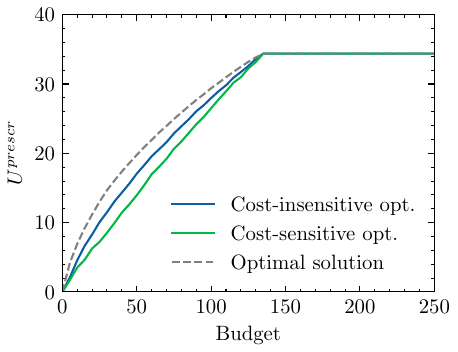}
        \caption{Cost-insensitive $\stackrel{}U^{presc}$ across increasing budget.}
        \label{fig:exp3_1_1}
    \end{subfigure}
    \begin{subfigure}[b]{0.35\textwidth}
        \includegraphics[width=\textwidth]{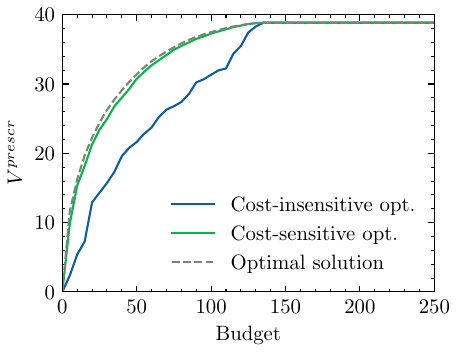}
        \caption{Cost-sensitive $\stackrel{}V^{presc}$ across increasing budget.}
        \label{fig:exp3_1_2}
    \end{subfigure}\\
    \caption{Comparison of cost-sensitive and cost-insensitive optimization, illustrating differences in performance based on $U^{presc}$ and $V^{presc}$. 
    The results underscore the importance of aligning the optimization objective with the downstream task.}
    \rule{\textwidth}{.5pt}
    \label{fig:exp3_1}
\end{figure}
Figure~\ref{fig:exp3_1_2} demonstrates that policies optimized with cost-sensitivity (green) perform better than cost-insensitive ones (blue) in terms of the cost-sensitive value ($V^{presc}$), particularly at lower budget levels. 
Conversely, Figure~\ref{fig:exp3_1_1} shows the opposite scenario, emphasizing the need to align the optimization process with the intended objective at hand.

This experiment highlights the importance of cost-sensitivity in optimizing treatment allocation under constrained budgets. By taking into account treatment costs and outcome benefits, cost-sensitive policies achieve higher values in cost-sensitive metrics, especially when resources are limited and not all entities can receive full treatment. These policies prioritize entities with the highest benefit-to-cost ratios, improving allocation efficiency. However, they underperform on cost-insensitive utility metrics, underlining the need for alignment between optimization objectives and specific decision-making goals.
\revised{%
The cost-sensitive policy, as expected, sacrifices some uplift if it leads to higher net value. We see differences in who gets treated: for example, some individuals with high uplift but low benefit may not be treated under the value-based policy, whereas they would under an uplift-only policy
}
These findings are relevant for applications such as healthcare, lending, and human resource management, where treatment costs and benefits vary across entities, regardless of whether the goal is to maximize profitability or allocate scarce resources more efficiently.
Furthermore, the modularity of the predict-then-optimize framework facilitates easy adaptation to changing objectives or constraints without requiring the retraining of predictive models. This makes it well-suited for dynamic environments where goals and resource availability evolve over time.
An extension of this experiment could explore the inclusion of application-dependent and entity-dependent treatment cost functions and their interaction with the previously researched fairness constraints, adding complexity to the optimization process.

\section{Conclusion, limitations, and further research}\label{sec:conclusion}

UM is extensively discussed in the literature, with applications such as lending, healthcare, HR, marketing, and maintenance. 
However, typically, no clear distinction between UM and CATE estimation is made, where CATE estimation should be viewed as a component of the broader UM field and is used as input for an optimization task.
By clearly distinguishing these steps, our approach accommodates more complex treatments, as demonstrated in our focus on continuous treatments, while allowing flexible integration of constraints like fairness considerations and cost-sensitive objectives.
Our main contributions are 
(i) defining UM, 
(ii) extending it to handle continuous treatments with customizable constraints and objectives, 
(iii) defining fairness considerations as explicit ILP constraints, and 
(iv) demonstrating its capabilities through experiments. 
This framework’s flexibility enables the use of pre-trained models and avoids challenges in decision-focused learning by allowing constraint integration without retraining, particularly advantageous when, for example, unexpected operational constraints arise. 
Continuous treatments allow fine-grained dose-based interventions, better suited for various applications, as shown in Table~\ref{tab:intro_applications}.

Our contributions are supported by the formal outlining and working of our framework in Sections~\ref{sec:uplift_modeling} and \ref{sec:problem}, as well as validation by a series of experiments in Section~\ref{sec:experiments}. 
Experiment 1 shows the benefits of continuous treatments in an uplift setting, enabling decision-makers to maximize the total benefit of treatment allocation. 
Interestingly, this experiment shows that the most accurate predictive models are not always the best suited for the downstream treatment allocation.
Experiment 2 explores the trade-offs between fairness constraints and policy value, revealing that stricter fairness enforcement, both in treatment allocation and outcomes, reduces overall utility, particularly when there are strong differences in ground-truth ATEs between the two groups considered. This reflects the broader challenge in algorithmic decision-making, where fairness may come at the cost of utility. 
Finally, Experiment 3 highlights the importance of aligning optimization with the specific objectives of the decision-making context, whether focusing on maximizing uplift or value.

While effective, our framework has limitations. 
\revised{
The ILP approach suffers from limited scalability. In contrast, the proposed heuristic handles larger problem sizes more efficiently. However, it may yield suboptimal solutions and does not account for side constraints such as fairness.
}
Additionally, the assumption of deterministic costs simplifies the problem but does not account for stochastic or uncertain treatment costs, which are present in other settings \citep{haupt2022targeting}.
Another limitation is the focus on single-phase treatments, where many real-world scenarios involve sequential decisions.
We also address fairness at the group level, leaving individual-level fairness for future exploration.
Finally, experiments are based on a single dataset and data-generating process, which may limit generalizability.

This work is the first step in defining and providing an initial solution for UM with continuous treatments.
Future research could extend this work by exploring varied allocation schemes, treatment characteristics, or sequential decision-making using dynamic policy or reinforcement learning.
In cases of time dependencies, treatments are administered multiple times over a period, or treatment effects change over time, 
dynamic policy learning or reinforcement learning methods could be employed to handle sequential treatments and long-term effects \citep{berrevoets2022treatment}.
Extending the framework to accommodate multiple treatment types, whether mutually exclusive or not, would create new opportunities in situations where multiple intervention types are available \citep{zhan2024weighted}.
Further research could investigate treatment effects in a network setting where spillover effects take place, making an entity's outcomes dependent on others' assigned doses and violating the stable unit treatment value assumption \citep{zhao2024learning}. 
Additionally, while this study focuses on one-dimensional dose-based treatments, other applications may involve a high-dimensional treatment space \citep{berrevoets2020organite}.
When optimizing under uncertainty, incorporating conformal predictions, where CATE distributions or intervals around CADRs are estimated instead of point estimates, could better inform personalized decision-making \citep{vanderschueren2023noflite,schroder2024conformal}. 
Finally, integrating DFL approaches, which, unlike our predict-then-optimize method, combine prediction and optimization in a single step, could potentially further improve policy outcomes by training models to directly make predictions that lead to better decisions \citep{mandi2023decision}.
\revised{
One could also explore hybrid approaches --– for example, incorporating fairness constraints directly into model training (a partial DFL approach) while still performing an optimization step for other parameters or constraints. Such methods might alleviate some of the burden on the post-prediction optimization
}

\section*{Acknowledgements}
This work was supported by Acerta Consult and the Flemish Government, through Flanders
Innovation \& Entrepreneurship (VLAIO, project HBC.2021.0833).

\section*{Declaration of generative AI and AI-assisted technologies in the writing process}

During the preparation of this work, the author(s) used ChatGPT and Grammarly in order to refine the phrasing of the manuscript. After using this tool/service, the author(s) reviewed and edited the content as needed and take(s) full responsibility for the content of the publication.



\clearpage

\appendix

\section{Notation overview}\label{app:notation}

\begin{table}[h]
    \centering
    \begin{adjustbox}{max width=1.0\textwidth}
    \begin{tabular}{cl}
    \toprule
         $\mathcal{D}=\left\{\left(\mathbf{x}_i, s_{i}, y_i\right)_{i=1}^{N}\right\}$  & Dataset containing $N$ entities\\
         $\mathcal{X}$                  & Feature space\\
         $\mathcal{S}$                  & Treatment space\\  
         $\mathcal{Y}$                  & Outcome space\\

         $\mathbf{X} \in \mathcal{X}$   & Features\\
         $S \in \mathcal{S}$            & Treatment\\
         $Y \in \mathcal{Y}$            & Outcome\\

         $\mu: \mathcal{S} \times \mathcal{X} \rightarrow [0,1]$ & Conditional average dose response function \\
         $\tau: \mathcal{S} \times \mathcal{X} \rightarrow [-1,1]$ & Conditional average dose effect function \\
         $\tau_s(\mathbf{x})$   &Treatment effect for the features-dose pair $\{\mathbf{x}, s\}$ \\

         $\hat{\mu}: \mathcal{S} \times \mathcal{X} \rightarrow [0,1]$ & Estimated conditional average dose response function \\
         $\hat{\tau}: \mathcal{S} \times \mathcal{X} \rightarrow [-1,1]$ & Estimated conditional average dose effect function \\
         $\hat{\tau}_s(\mathbf{x})$   &Estimated dose effect for the features-dose pair $\{\mathbf{x}, s\}$ \\
         
         {$\delta$}   &Number of dose bins \\
         $\hat{\bm{\tau}}(\mathbf{x}) \in [-1,1]^{\delta}$  & Vector with $\delta$ CADE estimates \\
         
         {$\pi: \mathcal{X} \rightarrow \{0,1\}^{\delta}$ }& Treatment assignment policy (unconstrained) for a single entity\\
         $\mathbf{C} \in \mathbb{R}^{(N \times \delta)}$ &Treatment cost matrix \\

         $\Psi_i(\pi) \in \mathbb{R}_{\geq0} $ &Treatment cost for a single entity \\
         $\mathbf{b} \in \mathbb{R}_{\geq0}^{N}$ &Benefit vector \\

         \revised{$\mathbf{T} \in [-1,1]^{(N\times\delta)}$} &\revised{Matrix with ($N\times\delta$) CADE values (ground-truth)}  \\
         \revised{$\mathbf{\hat{T}} \in [-1,1]^{(N\times\delta)}$}&\revised{Matrix with ($N\times\delta$) CADE values (estimated)}  \\
         
\cdashline{1-2}         
        \revised{
        $U_i: \{0,1\}^{\delta} \times [-1,1]^{\delta} \rightarrow \mathbb{R}$  
        }
        &\revised{\makecell[l]{ Policy uplift of single entity \\
                     Expected ($U_i^{exp}$), prescribed ($U_i^{presc}$), and optimal ($U_i^{opt}$)}} \\
                     
\cdashline{1-2}         $V_i: \{0,1\}^{\delta} \times [-1,1]^{\delta} \times \mathbb{R} \rightarrow \mathbb{R}$          &\makecell[l]{ Policy value of single entity \\
                     Expected ($V_i^{exp}$), prescribed ($V_i^{presc}$), and optimal ($V_i^{opt}$)} \\

\cdashline{1-2} 
         
         $\stackrel{B}\Pi: \mathbb{R^+}\times\mathcal{X}^N \rightarrow \{0,1\}^{(N\times\delta)}$ &\makecell[l]{Treatment assignment policy (constrained) over all entities \\ Expected optimal policy ($\stackrel{B}\Pi^{exp}$), ground-truth optimal policy ($\stackrel{B}\Pi^{opt}$)}    \\

\cdashline{1-2}
        \revised{
         $\stackrel{}U: \{0,1\}^{(N\times\delta)} \times [-1,1]^{(N\times\delta)} \rightarrow \mathbb{R}$}
         &
         \revised{\makecell[l]{ Policy uplift over all entities\\
                        Expected ($\stackrel{}U^{exp}$), prescribed ($\stackrel{}U^{presc}$), and optimal ($\stackrel{}U^{opt}$)} }\\

\cdashline{1-2}
    
         $\stackrel{}V: \{0,1\}^{(N\times\delta)} \times [-1,1]^{(N\times\delta)} \times \mathbb{R}^N \rightarrow \mathbb{R}$
         & \makecell[l]{ Policy value over all entities\\
                        Expected ($\stackrel{}V^{exp}$), prescribed ($\stackrel{}V^{presc}$), and optimal ($\stackrel{}V^{opt}$)} \\

\cdashline{1-2}

         $A$    &Protected sensitive attribute \\
         $\epsilon_{DT} \in [0,1]$       &Slack parameter for allocation fairness (Disparate Treatment)\\
         $\epsilon_{DO} \in [0,1]$    &Slack parameter for outcome fairness (Disparate Outcome) \\
    \bottomrule
    \end{tabular}
    \end{adjustbox}
\vspace{0.1cm}
    \caption{This table summarizes the notation in this paper.}
    \label{tab:notation_overview}
\end{table}

\clearpage
\section{Assumptions and mathematical justification of CADE identification}\label{app:assumptions}
\revised{
Under the Rubin-Neyman potential outcomes framework, the CADE \(\tau_s(\mathbf{x}) = \mathbb{E}[Y(s) - Y(0) \mid \mathbf{X} = \mathbf{x}]\) quantifies the expected outcome of a continuous treatment dose \(s\) relative to a baseline (i.e., no treatment). 
Based on the proof for the binary treatment setting by \citet{neal2020introduction}, we prove that $\tau_s(\mathbf{x})$ is identifiable as \eqref{eq:identified} and estimable via causal machine learning methods under three assumptions: Consistency, Ignorability, Overlap (Section~\ref{sec:methodology:cadr_estimation}).
}

\begin{proof}
From the definition of $\tau_s(\mathbf{x})$ and linearity of expectation:
\begin{align}
    \tau_s(\mathbf{x}) 
        &=\mathbb{E}[Y(s) - Y(0) \mid \mathbf{X} = \mathbf{x}] \\
        &= \mathbb{E}[Y(s) \mid \mathbf{X} = \mathbf{x}] - \mathbb{E}[Y(0) \mid \mathbf{X} = \mathbf{x}] \label{eq:cade_decomp}
\end{align}

By Ignorability, potential outcomes are independent of treatment assignment given $\mathbf{X}$:
\begin{align}
    \mathbb{E}[Y(s) \mid \mathbf{X} = \mathbf{x}] &= \mathbb{E}[Y(s) \mid \mathbf{X} = \mathbf{x}, S = s] \label{eq:ignorability_step}.
\end{align}

Applying Consistency to \eqref{eq:ignorability_step}:
\begin{align}
    \mathbb{E}[Y(s) \mid \mathbf{X} = \mathbf{x}, S = s] &= \mathbb{E}[Y \mid \mathbf{X} = \mathbf{x}, S = s]. \label{eq:consistency_step}
\end{align}

Combining \eqref{eq:cade_decomp}--\eqref{eq:consistency_step}:
\begin{align}
    \tau_s(\mathbf{x}) &= \mathbb{E}[Y \mid \mathbf{X} = \mathbf{x}, S = s] - \mathbb{E}[Y \mid \mathbf{X} = \mathbf{x}, S = 0], \label{eq:identified}
\end{align}
where Overlap ensures the conditional expectations in \eqref{eq:ignorability_step}-\eqref{eq:identified} are well-defined.
\end{proof}


\section{Details regarding semi-synthetic data}\label{app:ihdp}

The original IHDP dataset contains 747 observations and 25 features. Following \citet{nie2021vcnet}, synthetic counterfactuals are generated by Equations~\ref{eq:ihdp_t_tilde}-\ref{eq:ihdp_y} where 
$S_{\mathrm{con}} = \bigl\{ 1, 2, 3, 5, 6 \bigr\} $ is the index set of continuous features,
$S_{\mathrm{bin}, 1} = \bigl\{ 4, 7, 8, 9, 10, 11, 12, 13, 14, 15\bigr\}$
and $S_{\mathrm{bin}, 2} = \bigl\{ 16, 17, 18, 19, 20, 21, 22, 23, 24, 25\bigr\}$ are two sets of binary features,
$c_1=\mathbb{E}\left[\frac{\sum_{i \in S_{\mathrm{dis}, 1}} x_i}{|S_{\mathrm{bin}, 1}|}\right]$, 
and $c_2=\mathbb{E}\left[\frac{\sum_{i \in S_{\mathrm{dis}, 2}} x_i}{|S_{\mathrm{bin}, 2}|}\right]$.

\begin{equation}\label{eq:ihdp_t_tilde}
    \tilde{t} \mid \boldsymbol{x} =
    \frac{x_1}{\left(1+x_2\right)}+\frac{\max \left(x_3, x_5, x_6\right)}{0.2+\min \left(x_3, x_5, x_6\right)}+\tanh \left(5 \frac{\sum_{i \in S_{\mathrm{bin}, 2}}\left(x_i-c_2\right)}{\left|S_{\mathrm{bin}, 2}\right|}\right)-2+\mathcal{N}(0,0.25)
\end{equation}

\begin{equation}\label{eq:ihdp_t}
    t =
    \bigl(1+\exp(-2\tilde{t})\bigr)^{-1}
\end{equation}

\begin{equation}\label{eq:ihdp_y_tilde}
    \tilde{y} \mid \boldsymbol{x},t =
    \frac{\sin (3 \pi t)}{1.2-t}\left(\tanh \left(5 \frac{\sum_{i \in S_{\mathrm{bin}, 1}}\left(x_i-c_1\right)}{\left|S_{\mathrm{bin}, 1}\right|}\right)+\frac{\exp \left(0.2\left(x_1-x_6\right)\right)}{0.5+5 \min \left(x_2, x_3, x_5\right)}\right)+\mathcal{N}(0,0.25)
\end{equation}

\begin{equation}\label{eq:ihdp_y}
    y =
    \frac{\tilde{y}-\min(\tilde{{y}})}{\max(\tilde{{y})}-\min(\tilde{{y}})}
\end{equation}

Following the literature, features are preprocessed to follow a standard normal distribution, and the generated treatments are normalized to fall within the range [0, 1] \citep{hill2011bayesian,nie2021vcnet}.
Additionally, we also standardize the outcomes so that $y\in [0,1]$.

\clearpage
\section{Number of dose bins $\delta$}\label{app:delta}

\begin{figure}[h]
    \centering
    \includegraphics[width=0.55\linewidth]{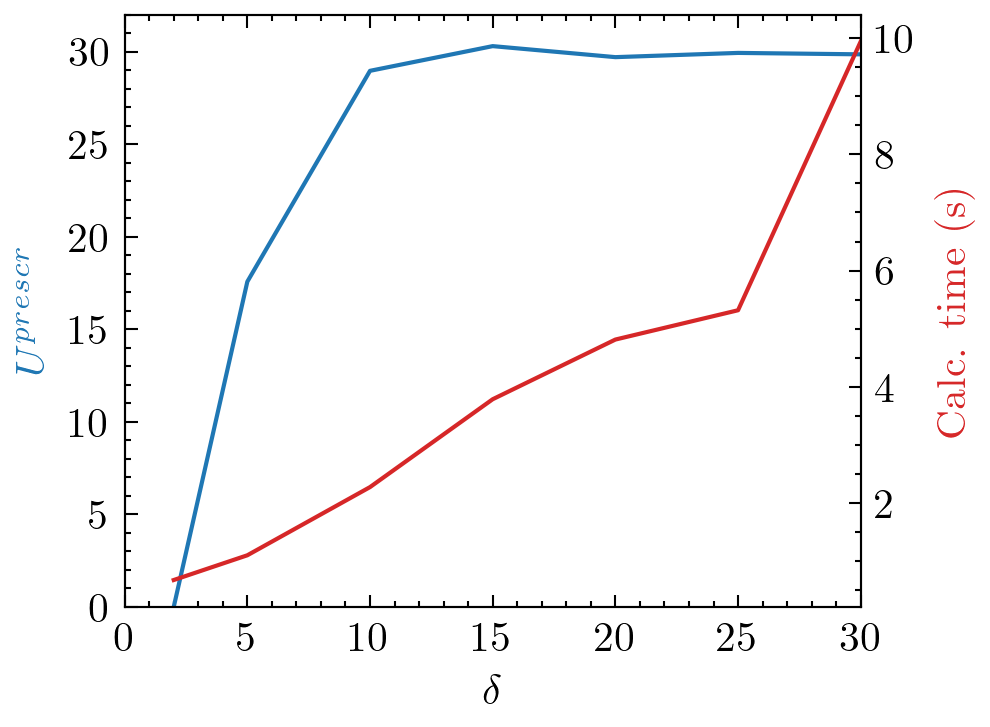}
    \caption{Effect of available dose bins on $U^{presc}$ and calculation time}
    \label{fig:exp_delta}
\end{figure}

Throughout the experiments, the number of available treatment bins $\delta$ is set to 10.
Figure~\ref{fig:exp_delta} illustrates the relationship between the number of treatment bins $\delta$, $U^{presc}$, and calculation time in seconds. 
For the semi-synthetic IHDP dataset, the increased benefit of using smaller-grained bins quickly caps out, while the required computation time continues to rise.
From an application perspective, also not each granularity of doses should be considered.
For instance, lending costs occur only in specific increments, and in an HR setting, training hours are typically scheduled in multiples of the session duration.
Given these considerations, $\delta$ is set to 10.

\clearpage
\section{Hyperparameter tuning}\label{app:hyperparameters}

    \begin{table}[h]
    \centering
        \begin{adjustbox}{max width=1.0\textwidth}
        
        \begin{tabular}{lllc}
        
        \toprule 

        \textbf{\textit{Method}}                & \textbf{Hyperparameter}    & \textbf{Values}                                            & \textbf{\#Models}               \\
        
        \midrule
        
        \multirow{4}{*}{\textit{S-Learner (rf)}}   
                & Estimators        & \{10, 20, 50, 100, \underline{200}, 500\}                          & \multirow{4}{*}{72}   \\
                & Criterion         & \{\textit{Sq. error}, \underline{\textit{Abs. error}}\}   &                       \\
                & Max depth         & \{5, \underline{15}, \textit{None}\}      &                       \\
                & Max features      & \{\underline{\textit{Sqrt}}, \textit{Log2}\}                &                       \\
                              
        \midrule

        \multirow{7}{*}{\textit{S-Learner (mlp)}}  
                & Learning rate     & \{0.001, \underline{0.01}\}     & \multirow{7}{*}{128}  \\
                & L2 regularization & \{\underline{0.0}, 0.1\}        &                       \\
                & Batch size        & \{\underline{64}, 128\}          &                       \\
                & Num layers        & \{2, \underline{3}\}              &                       \\
                & Hidden size       & \{\underline{32}, 64\}            &                       \\
                & Steps             & \{500, 1000, 2000, \underline{5000}\}              &                       \\
                & Optimizer         & \{\textit{Adam}\}              & \\
       
        \midrule

        \multirow{9}{*}{\textit{DRNet}}
                & Learning rate     & \{\underline{0.001}, 0.01\}  & \multirow{9}{*}{256}   \\
                & L2 regularization & \{\underline{0.0}, 0.1\}        &                       \\
                & Batch size        & \{64, \underline{128}\}          &                       \\
                & Num repr. layers        & \{2, \underline{3}\}              &                       \\
                & Num inf. layers        & \{\underline{1}, 2\}              &                   \\
                & Hidden size       & \{32, \underline{64}\}            &                       \\
                & Steps             & \{500, 1000, 2000, \underline{5000}\}              &                       \\
                & Num dose strata       & \{10\}              &   \\
                
                & Optimizer         & \{\textit{Adam}\} & \\

        \midrule
                
        \multirow{5}{*}{\textit{VCNet}}
                & Learning rate     & \{0.001, \underline{0.01}\}   & \multirow{5}{*}{32}  \\
                & Batch size        & \{64, \underline{128}\}          &                       \\
                & Hidden size       & \{32, \underline{64}\}            &                       \\
                & Steps             & \{500, 1000, \underline{2000}, 5000\}              &                       \\
                & Optimizer         & \{\textit{Adam}\} & \\


        \bottomrule

        \end{tabular}
        \end{adjustbox}
\vspace{0.1cm}
\caption{Hyperparameter search space. The selected hyperparameters are underlined.}

        \label{tab:hyperpara}
    \end{table}

\clearpage

\section{More details on Experiment 1}\label{app:exp1}

\subsection{Dose-response estimation}\label{app:exp1_cadr}

\begin{figure}[h]
    \centering
        \begin{subfigure}[b]{0.49\textwidth}
        \includegraphics[width=\textwidth]{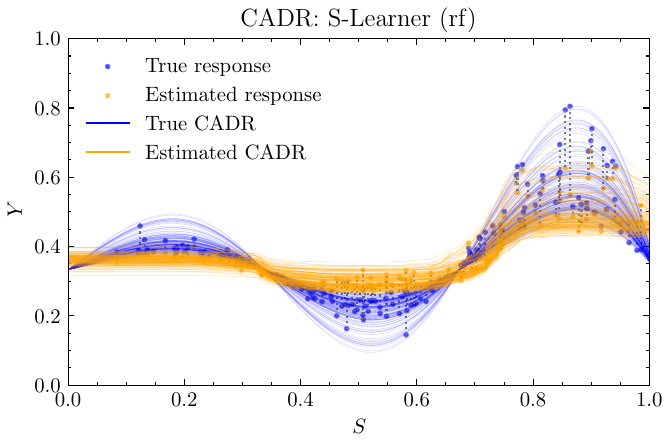}
        \caption{
        }
        \label{fig:app_exp1_cadr_slearner}
    \end{subfigure}
    \hfill
    \begin{subfigure}[b]{0.49\textwidth}
        \includegraphics[width=\textwidth]{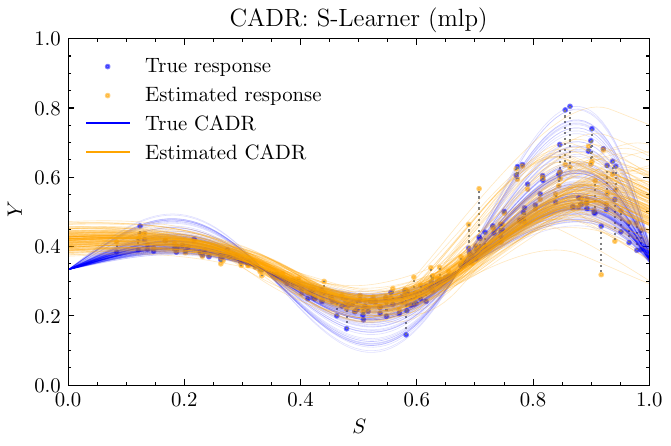}
        \caption{
        }
        \label{fig:app_exp1_cadr_mlp}
    \end{subfigure}

    \begin{subfigure}[b]{0.49\textwidth}
        \includegraphics[width=\textwidth]{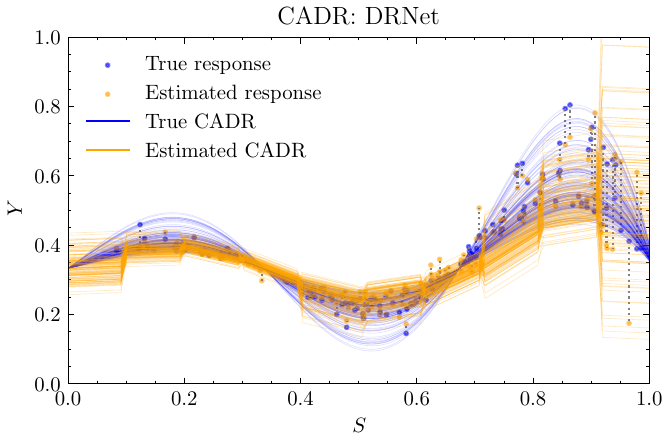}
        \caption{
        }
        \label{fig:app_exp1_cadr_drnet}
    \end{subfigure}
    \hfill
    \begin{subfigure}[b]{0.49\textwidth}
        \includegraphics[width=\textwidth]{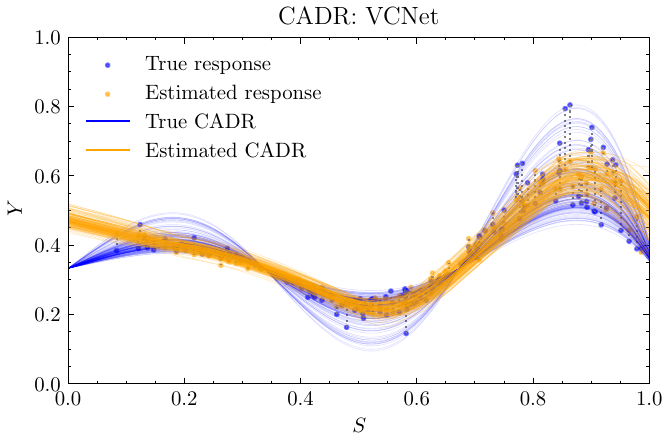}
        \caption{
        }
        \label{fig:app_exp1_cadr_vcnet}
    \end{subfigure}
    \caption{This figure presents the ground-truth dose responses in blue, with the blue dots representing the factual outcomes, and their corresponding estimates in orange, for four different predictive methods on a separate test set from the IHDP dataset.}
    \label{fig:subplots}
\end{figure}

\clearpage
\subsection{\revised{Scalability}}\label{app:exp1_scalability}

\begin{figure}[h]
    \centering
    \includegraphics[width=0.5\linewidth]{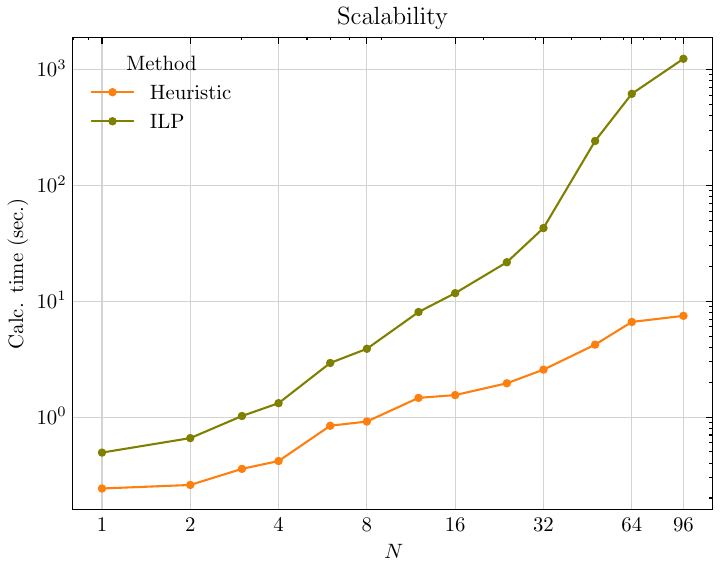}
    \caption{Runtime of the ILP and heuristic as a function of problem size. The original dataset size corresponds with $N=1$.}
    \label{fig:app_exp1_scalability}
\end{figure}

\revised{
We provide additional runtime analysis for increasing dataset sizes. The dataset size is semi-synthetically scaled by a factor \(N\) (with \(N = 1\) corresponding to the original dataset), by randomly oversampling observations and adding slight noise to ensure that each observation remains unique.
We adopt the setup of Experiment~1 (Section~\ref{sec:exp1}), where the available budget is also scaled by $N$, and compare the runtime of the ILP solution with that of the heuristic approach. Figure~\ref{fig:app_exp1_scalability} shows the runtime (in seconds) as a function of the problem size (measured by the scaling factor \(N\)). Both axes are shown on a logarithmic scale.
The clear advantage of the heuristic method is that it scales well. However, it does not take into account any side constraints, which are allowed by the ILP. The downside of the ILP is that runtimes 
increase rapidly with the problem size, becoming impractical for large-scale instances. This highlights the classic trade-off: while the ILP allows for more expressive modeling and constraint handling, its computational cost means that problem size is limited. In contrast, the heuristic remains feasible in terms of computation time even as the dataset grows.
}

\clearpage

\bibliographystyle{elsarticle-harv} 
\bibliography{bibliography}

\begin{thebibliography}{73}
\expandafter\ifx\csname natexlab\endcsname\relax\def\natexlab#1{#1}\fi
\providecommand{\url}[1]{\texttt{#1}}
\providecommand{\href}[2]{#2}
\providecommand{\path}[1]{#1}
\providecommand{\DOIprefix}{doi:}
\providecommand{\ArXivprefix}{arXiv:}
\providecommand{\URLprefix}{URL: }
\providecommand{\Pubmedprefix}{pmid:}
\providecommand{\doi}[1]{\href{http://dx.doi.org/#1}{\path{#1}}}
\providecommand{\Pubmed}[1]{\href{pmid:#1}{\path{#1}}}
\providecommand{\bibinfo}[2]{#2}
\ifx\xfnm\relax \def\xfnm[#1]{\unskip,\space#1}\fi
\bibitem[{Bahnsen et~al.(2014)Bahnsen, Aouada and Ottersten}]{bahnsen2014example}
\bibinfo{author}{Bahnsen, A.C.}, \bibinfo{author}{Aouada, D.}, \bibinfo{author}{Ottersten, B.}, \bibinfo{year}{2014}.
\newblock \bibinfo{title}{Example-dependent cost-sensitive logistic regression for credit scoring}, in: \bibinfo{booktitle}{2014 13th International conference on machine learning and applications}, \bibinfo{organization}{IEEE}. pp. \bibinfo{pages}{263--269}.
\bibitem[{Barocas et~al.(2023)Barocas, Hardt and Narayanan}]{barocas2023fairness}
\bibinfo{author}{Barocas, S.}, \bibinfo{author}{Hardt, M.}, \bibinfo{author}{Narayanan, A.}, \bibinfo{year}{2023}.
\newblock \bibinfo{title}{Fairness and machine learning: Limitations and opportunities}.
\newblock \bibinfo{publisher}{MIT Press}.
\bibitem[{Berrevoets et~al.(2020)Berrevoets, Jordon, Bica, van~der Schaar et~al.}]{berrevoets2020organite}
\bibinfo{author}{Berrevoets, J.}, \bibinfo{author}{Jordon, J.}, \bibinfo{author}{Bica, I.}, \bibinfo{author}{van~der Schaar, M.}, et~al., \bibinfo{year}{2020}.
\newblock \bibinfo{title}{Organite: Optimal transplant donor organ offering using an individual treatment effect}.
\newblock \bibinfo{journal}{Advances in neural information processing systems} \bibinfo{volume}{33}, \bibinfo{pages}{20037--20050}.
\bibitem[{Berrevoets et~al.(2022)Berrevoets, Verboven and Verbeke}]{berrevoets2022treatment}
\bibinfo{author}{Berrevoets, J.}, \bibinfo{author}{Verboven, S.}, \bibinfo{author}{Verbeke, W.}, \bibinfo{year}{2022}.
\newblock \bibinfo{title}{Treatment effect optimisation in dynamic environments}.
\newblock \bibinfo{journal}{Journal of Causal Inference} \bibinfo{volume}{10}, \bibinfo{pages}{106--122}.
\bibitem[{Betlei et~al.(2021)Betlei, Diemert and Amini}]{betlei2021uplift}
\bibinfo{author}{Betlei, A.}, \bibinfo{author}{Diemert, E.}, \bibinfo{author}{Amini, M.R.}, \bibinfo{year}{2021}.
\newblock \bibinfo{title}{Uplift modeling with generalization guarantees}, in: \bibinfo{booktitle}{Proceedings of the 27th ACM SIGKDD Conference on Knowledge Discovery \& Data Mining}, pp. \bibinfo{pages}{55--65}.
\bibitem[{Bica et~al.(2020)Bica, Jordon and van~der Schaar}]{bica2020estimating}
\bibinfo{author}{Bica, I.}, \bibinfo{author}{Jordon, J.}, \bibinfo{author}{van~der Schaar, M.}, \bibinfo{year}{2020}.
\newblock \bibinfo{title}{Estimating the effects of continuous-valued interventions using generative adversarial networks}.
\newblock \bibinfo{journal}{Advances in Neural Information Processing Systems} \bibinfo{volume}{33}, \bibinfo{pages}{16434--16445}.
\bibitem[{Bockel-Rickermann et~al.(2023)Bockel-Rickermann, Vanderschueren, Berrevoets, Verdonck and Verbeke}]{bockel2023learning}
\bibinfo{author}{Bockel-Rickermann, C.}, \bibinfo{author}{Vanderschueren, T.}, \bibinfo{author}{Berrevoets, J.}, \bibinfo{author}{Verdonck, T.}, \bibinfo{author}{Verbeke, W.}, \bibinfo{year}{2023}.
\newblock \bibinfo{title}{Learning continuous-valued treatment effects through representation balancing}.
\newblock \bibinfo{journal}{arXiv preprint arXiv:2309.03731} .
\bibitem[{Bockel-Rickermann et~al.(2024)Bockel-Rickermann, Vanderschueren, Verdonck and Verbeke}]{bockel2024sources}
\bibinfo{author}{Bockel-Rickermann, C.}, \bibinfo{author}{Vanderschueren, T.}, \bibinfo{author}{Verdonck, T.}, \bibinfo{author}{Verbeke, W.}, \bibinfo{year}{2024}.
\newblock \bibinfo{title}{Sources of gain: Decomposing performance in conditional average dose response estimation}.
\newblock \bibinfo{journal}{arXiv preprint arXiv:2406.08206} .
\bibitem[{Brooks-Gunn et~al.(1992)Brooks-Gunn, Liaw and Klebanov}]{brooks1992effects}
\bibinfo{author}{Brooks-Gunn, J.}, \bibinfo{author}{Liaw, F.r.}, \bibinfo{author}{Klebanov, P.K.}, \bibinfo{year}{1992}.
\newblock \bibinfo{title}{Effects of early intervention on cognitive function of low birth weight preterm infants}.
\newblock \bibinfo{journal}{The Journal of pediatrics} \bibinfo{volume}{120}, \bibinfo{pages}{350--359}.
\bibitem[{Chouldechova and Roth(2020)}]{chouldechova2020snapshot}
\bibinfo{author}{Chouldechova, A.}, \bibinfo{author}{Roth, A.}, \bibinfo{year}{2020}.
\newblock \bibinfo{title}{A snapshot of the frontiers of fairness in machine learning}.
\newblock \bibinfo{journal}{Communications of the ACM} \bibinfo{volume}{63}, \bibinfo{pages}{82--89}.
\bibitem[{Corbett-Davies et~al.(2017)Corbett-Davies, Pierson, Feller, Goel and Huq}]{corbett2017algorithmic}
\bibinfo{author}{Corbett-Davies, S.}, \bibinfo{author}{Pierson, E.}, \bibinfo{author}{Feller, A.}, \bibinfo{author}{Goel, S.}, \bibinfo{author}{Huq, A.}, \bibinfo{year}{2017}.
\newblock \bibinfo{title}{Algorithmic decision making and the cost of fairness}, in: \bibinfo{booktitle}{Proceedings of the 23rd acm sigkdd international conference on knowledge discovery and data mining}, pp. \bibinfo{pages}{797--806}.
\bibitem[{Dantzig(1957)}]{dantzig1957discrete}
\bibinfo{author}{Dantzig, G.B.}, \bibinfo{year}{1957}.
\newblock \bibinfo{title}{Discrete-variable extremum problems}.
\newblock \bibinfo{journal}{Operations Research} \bibinfo{volume}{5}, \bibinfo{pages}{266--288}.
\bibitem[{Dastin(2022)}]{dastin2022amazon}
\bibinfo{author}{Dastin, J.}, \bibinfo{year}{2022}.
\newblock \bibinfo{title}{Amazon scraps secret ai recruiting tool that showed bias against women}, in: \bibinfo{booktitle}{Ethics of data and analytics}. \bibinfo{publisher}{Auerbach Publications}, pp. \bibinfo{pages}{296--299}.
\bibitem[{De~Bock et~al.(2020)De~Bock, Coussement and Lessmann}]{de2020cost}
\bibinfo{author}{De~Bock, K.W.}, \bibinfo{author}{Coussement, K.}, \bibinfo{author}{Lessmann, S.}, \bibinfo{year}{2020}.
\newblock \bibinfo{title}{Cost-sensitive business failure prediction when misclassification costs are uncertain: A heterogeneous ensemble selection approach}.
\newblock \bibinfo{journal}{European Journal of Operational Research} \bibinfo{volume}{285}, \bibinfo{pages}{612--630}.
\bibitem[{De~Vos et~al.(2023)De~Vos, Vanderschueren, Verdonck and Verbeke}]{de2023robust}
\bibinfo{author}{De~Vos, S.}, \bibinfo{author}{Vanderschueren, T.}, \bibinfo{author}{Verdonck, T.}, \bibinfo{author}{Verbeke, W.}, \bibinfo{year}{2023}.
\newblock \bibinfo{title}{Robust instance-dependent cost-sensitive classification}.
\newblock \bibinfo{journal}{Advances in Data Analysis and Classification} \bibinfo{volume}{17}, \bibinfo{pages}{1057--1079}.
\bibitem[{Devriendt et~al.(2021)Devriendt, Berrevoets and Verbeke}]{devriendt2021you}
\bibinfo{author}{Devriendt, F.}, \bibinfo{author}{Berrevoets, J.}, \bibinfo{author}{Verbeke, W.}, \bibinfo{year}{2021}.
\newblock \bibinfo{title}{Why you should stop predicting customer churn and start using uplift models}.
\newblock \bibinfo{journal}{Information Sciences} \bibinfo{volume}{548}, \bibinfo{pages}{497--515}.
\bibitem[{Devriendt et~al.(2018)Devriendt, Moldovan and Verbeke}]{devriendt2018literature}
\bibinfo{author}{Devriendt, F.}, \bibinfo{author}{Moldovan, D.}, \bibinfo{author}{Verbeke, W.}, \bibinfo{year}{2018}.
\newblock \bibinfo{title}{A literature survey and experimental evaluation of the state-of-the-art in uplift modeling: A stepping stone toward the development of prescriptive analytics}.
\newblock \bibinfo{journal}{Big Data} \bibinfo{volume}{6}, \bibinfo{pages}{13--41}.
\bibitem[{Devriendt et~al.(2020)Devriendt, Van~Belle, Guns and Verbeke}]{devriendt2020learning}
\bibinfo{author}{Devriendt, F.}, \bibinfo{author}{Van~Belle, J.}, \bibinfo{author}{Guns, T.}, \bibinfo{author}{Verbeke, W.}, \bibinfo{year}{2020}.
\newblock \bibinfo{title}{Learning to rank for uplift modeling}.
\newblock \bibinfo{journal}{IEEE Transactions on Knowledge and Data Engineering} \bibinfo{volume}{34}, \bibinfo{pages}{4888--4904}.
\bibitem[{Dressel and Farid(2018)}]{dressel2018accuracy}
\bibinfo{author}{Dressel, J.}, \bibinfo{author}{Farid, H.}, \bibinfo{year}{2018}.
\newblock \bibinfo{title}{The accuracy, fairness, and limits of predicting recidivism}.
\newblock \bibinfo{journal}{Science advances} \bibinfo{volume}{4}, \bibinfo{pages}{eaao5580}.
\bibitem[{Elmachtoub and Grigas(2022)}]{elmachtoub2022smart}
\bibinfo{author}{Elmachtoub, A.N.}, \bibinfo{author}{Grigas, P.}, \bibinfo{year}{2022}.
\newblock \bibinfo{title}{Smart “predict, then optimize”}.
\newblock \bibinfo{journal}{Management Science} \bibinfo{volume}{68}, \bibinfo{pages}{9--26}.
\bibitem[{{European Commission}(2021)}]{european_ai_act_2021}
\bibinfo{author}{{European Commission}}, \bibinfo{year}{2021}.
\newblock \bibinfo{title}{{Proposal for a Regulation of the European Parliament and of the Council Laying Down Harmonised Rules on Artificial Intelligence (Artificial Intelligence Act) and Amending Certain Union Legislative Acts}}.
\newblock \bibinfo{note}{COM(2021) 206 final}.
\bibitem[{Fern{\'a}ndez-Lor{\'\i}a and Provost(2022)}]{fernandez2022causal}
\bibinfo{author}{Fern{\'a}ndez-Lor{\'\i}a, C.}, \bibinfo{author}{Provost, F.}, \bibinfo{year}{2022}.
\newblock \bibinfo{title}{Causal decision making and causal effect estimation are not the same… and why it matters}.
\newblock \bibinfo{journal}{INFORMS Journal on Data Science} \bibinfo{volume}{1}, \bibinfo{pages}{4--16}.
\bibitem[{Frauen et~al.(2023)Frauen, Melnychuk and Feuerriegel}]{frauen2023fair}
\bibinfo{author}{Frauen, D.}, \bibinfo{author}{Melnychuk, V.}, \bibinfo{author}{Feuerriegel, S.}, \bibinfo{year}{2023}.
\newblock \bibinfo{title}{Fair off-policy learning from observational data}.
\newblock \bibinfo{journal}{arXiv preprint arXiv:2303.08516} .
\bibitem[{Gubela and Lessmann(2021)}]{gubela2021uplift}
\bibinfo{author}{Gubela, R.M.}, \bibinfo{author}{Lessmann, S.}, \bibinfo{year}{2021}.
\newblock \bibinfo{title}{Uplift modeling with value-driven evaluation metrics}.
\newblock \bibinfo{journal}{Decision Support Systems} \bibinfo{volume}{150}, \bibinfo{pages}{113648}.
\bibitem[{Gubela et~al.(2020)Gubela, Lessmann and Jaroszewicz}]{gubela2020response}
\bibinfo{author}{Gubela, R.M.}, \bibinfo{author}{Lessmann, S.}, \bibinfo{author}{Jaroszewicz, S.}, \bibinfo{year}{2020}.
\newblock \bibinfo{title}{Response transformation and profit decomposition for revenue uplift modeling}.
\newblock \bibinfo{journal}{European Journal of Operational Research} \bibinfo{volume}{283}, \bibinfo{pages}{647--661}.
\bibitem[{Guo et~al.(2020)Guo, Cheng, Li, Hahn and Liu}]{guo2020survey}
\bibinfo{author}{Guo, R.}, \bibinfo{author}{Cheng, L.}, \bibinfo{author}{Li, J.}, \bibinfo{author}{Hahn, P.R.}, \bibinfo{author}{Liu, H.}, \bibinfo{year}{2020}.
\newblock \bibinfo{title}{A survey of learning causality with data: Problems and methods}.
\newblock \bibinfo{journal}{ACM Computing Surveys (CSUR)} \bibinfo{volume}{53}, \bibinfo{pages}{1--37}.
\bibitem[{Gutierrez and G{\'e}rardy(2017)}]{gutierrez2017causal}
\bibinfo{author}{Gutierrez, P.}, \bibinfo{author}{G{\'e}rardy, J.Y.}, \bibinfo{year}{2017}.
\newblock \bibinfo{title}{Causal inference and uplift modelling: A review of the literature}, in: \bibinfo{booktitle}{International conference on predictive applications and APIs}, \bibinfo{organization}{PMLR}. pp. \bibinfo{pages}{1--13}.
\bibitem[{Han et~al.(2023)Han, Jiang, Jin, Liu, Zou, Wang and Hu}]{han2023retiring}
\bibinfo{author}{Han, X.}, \bibinfo{author}{Jiang, Z.}, \bibinfo{author}{Jin, H.}, \bibinfo{author}{Liu, Z.}, \bibinfo{author}{Zou, N.}, \bibinfo{author}{Wang, Q.}, \bibinfo{author}{Hu, X.}, \bibinfo{year}{2023}.
\newblock \bibinfo{title}{Retiring deltadp: New distribution-level metrics for demographic parity}.
\newblock \bibinfo{journal}{Transactions on Machine Learning Research} .
\bibitem[{Hatt and Feuerriegel(2024)}]{hatt2024sequential}
\bibinfo{author}{Hatt, T.}, \bibinfo{author}{Feuerriegel, S.}, \bibinfo{year}{2024}.
\newblock \bibinfo{title}{Sequential deconfounding for causal inference with unobserved confounders}, in: \bibinfo{booktitle}{Causal Learning and Reasoning}, \bibinfo{organization}{PMLR}. pp. \bibinfo{pages}{934--956}.
\bibitem[{Haupt and Lessmann(2022)}]{haupt2022targeting}
\bibinfo{author}{Haupt, J.}, \bibinfo{author}{Lessmann, S.}, \bibinfo{year}{2022}.
\newblock \bibinfo{title}{Targeting customers under response-dependent costs}.
\newblock \bibinfo{journal}{European Journal of Operational Research} \bibinfo{volume}{297}, \bibinfo{pages}{369--379}.
\bibitem[{Hill(2011)}]{hill2011bayesian}
\bibinfo{author}{Hill, J.L.}, \bibinfo{year}{2011}.
\newblock \bibinfo{title}{Bayesian nonparametric modeling for causal inference}.
\newblock \bibinfo{journal}{Journal of Computational and Graphical Statistics} \bibinfo{volume}{20}, \bibinfo{pages}{217--240}.
\bibitem[{Hirano and Imbens(2004)}]{hirano2004propensity}
\bibinfo{author}{Hirano, K.}, \bibinfo{author}{Imbens, G.}, \bibinfo{year}{2004}.
\newblock \bibinfo{title}{The propensity score with continuous treatments. applied bayesian modeling and causal inference from incomplete-data perspectives}.
\bibitem[{Holland(1986)}]{holland1986statistics}
\bibinfo{author}{Holland, P.W.}, \bibinfo{year}{1986}.
\newblock \bibinfo{title}{Statistics and causal inference}.
\newblock \bibinfo{journal}{Journal of the American statistical Association} \bibinfo{volume}{81}, \bibinfo{pages}{945--960}.
\bibitem[{Holland-Letz and Kopp-Schneider(2015)}]{holland2015optimal}
\bibinfo{author}{Holland-Letz, T.}, \bibinfo{author}{Kopp-Schneider, A.}, \bibinfo{year}{2015}.
\newblock \bibinfo{title}{Optimal experimental designs for dose--response studies with continuous endpoints}.
\newblock \bibinfo{journal}{Archives of toxicology} \bibinfo{volume}{89}, \bibinfo{pages}{2059--2068}.
\bibitem[{H{\"o}ppner et~al.(2022)H{\"o}ppner, Baesens, Verbeke and Verdonck}]{hoppner2022instance}
\bibinfo{author}{H{\"o}ppner, S.}, \bibinfo{author}{Baesens, B.}, \bibinfo{author}{Verbeke, W.}, \bibinfo{author}{Verdonck, T.}, \bibinfo{year}{2022}.
\newblock \bibinfo{title}{Instance-dependent cost-sensitive learning for detecting transfer fraud}.
\newblock \bibinfo{journal}{European Journal of Operational Research} \bibinfo{volume}{297}, \bibinfo{pages}{291--300}.
\bibitem[{Imbens(2000)}]{imbens2000role}
\bibinfo{author}{Imbens, G.W.}, \bibinfo{year}{2000}.
\newblock \bibinfo{title}{The role of the propensity score in estimating dose-response functions}.
\newblock \bibinfo{journal}{Biometrika} \bibinfo{volume}{87}, \bibinfo{pages}{706--710}.
\bibitem[{Jaskowski and Jaroszewicz(2012)}]{jaskowski2012uplift}
\bibinfo{author}{Jaskowski, M.}, \bibinfo{author}{Jaroszewicz, S.}, \bibinfo{year}{2012}.
\newblock \bibinfo{title}{Uplift modeling for clinical trial data}, in: \bibinfo{booktitle}{ICML workshop on clinical data analysis}, pp. \bibinfo{pages}{79--95}.
\bibitem[{Johansson et~al.(2016)Johansson, Shalit and Sontag}]{johansson2016learning}
\bibinfo{author}{Johansson, F.}, \bibinfo{author}{Shalit, U.}, \bibinfo{author}{Sontag, D.}, \bibinfo{year}{2016}.
\newblock \bibinfo{title}{Learning representations for counterfactual inference}, in: \bibinfo{booktitle}{International conference on machine learning}, \bibinfo{organization}{PMLR}. pp. \bibinfo{pages}{3020--3029}.
\bibitem[{Jones et~al.(2019)Jones, Molitor and Reif}]{jones2019workplace}
\bibinfo{author}{Jones, D.}, \bibinfo{author}{Molitor, D.}, \bibinfo{author}{Reif, J.}, \bibinfo{year}{2019}.
\newblock \bibinfo{title}{What do workplace wellness programs do? evidence from the illinois workplace wellness study}.
\newblock \bibinfo{journal}{The Quarterly Journal of Economics} \bibinfo{volume}{134}, \bibinfo{pages}{1747--1791}.
\bibitem[{Khan et~al.(2023)Khan, Herasymuk and Stoyanovich}]{khan2023fairness}
\bibinfo{author}{Khan, F.A.}, \bibinfo{author}{Herasymuk, D.}, \bibinfo{author}{Stoyanovich, J.}, \bibinfo{year}{2023}.
\newblock \bibinfo{title}{On fairness and stability: Is estimator variance a friend or a foe?}
\newblock \bibinfo{journal}{arXiv preprint: 2302.04525} .
\bibitem[{Kleinberg et~al.(2016)Kleinberg, Mullainathan and Raghavan}]{kleinberg2016inherent}
\bibinfo{author}{Kleinberg, J.}, \bibinfo{author}{Mullainathan, S.}, \bibinfo{author}{Raghavan, M.}, \bibinfo{year}{2016}.
\newblock \bibinfo{title}{Inherent trade-offs in the fair determination of risk scores}.
\newblock \bibinfo{journal}{arXiv preprint arXiv:1609.05807} .
\bibitem[{Kozodoi et~al.(2022)Kozodoi, Jacob and Lessmann}]{kozodoi2022fairness}
\bibinfo{author}{Kozodoi, N.}, \bibinfo{author}{Jacob, J.}, \bibinfo{author}{Lessmann, S.}, \bibinfo{year}{2022}.
\newblock \bibinfo{title}{Fairness in credit scoring: Assessment, implementation and profit implications}.
\newblock \bibinfo{journal}{European Journal of Operational Research} \bibinfo{volume}{297}, \bibinfo{pages}{1083--1094}.
\bibitem[{K{\"u}nzel et~al.(2019)K{\"u}nzel, Sekhon, Bickel and Yu}]{kunzel2019metalearners}
\bibinfo{author}{K{\"u}nzel, S.R.}, \bibinfo{author}{Sekhon, J.S.}, \bibinfo{author}{Bickel, P.J.}, \bibinfo{author}{Yu, B.}, \bibinfo{year}{2019}.
\newblock \bibinfo{title}{Metalearners for estimating heterogeneous treatment effects using machine learning}.
\newblock \bibinfo{journal}{Proceedings of the national academy of sciences} \bibinfo{volume}{116}, \bibinfo{pages}{4156--4165}.
\bibitem[{Lechner(2001)}]{lechner2001identification}
\bibinfo{author}{Lechner, M.}, \bibinfo{year}{2001}.
\newblock \bibinfo{title}{Identification and estimation of causal effects of multiple treatments under the conditional independence assumption}.
\newblock \bibinfo{publisher}{Springer}.
\bibitem[{Lemmens and Gupta(2020)}]{lemmens2020managing}
\bibinfo{author}{Lemmens, A.}, \bibinfo{author}{Gupta, S.}, \bibinfo{year}{2020}.
\newblock \bibinfo{title}{Managing churn to maximize profits}.
\newblock \bibinfo{journal}{Marketing Science} \bibinfo{volume}{39}, \bibinfo{pages}{956--973}.
\bibitem[{Makhlouf et~al.(2021)Makhlouf, Zhioua and Palamidessi}]{makhlouf2021applicability}
\bibinfo{author}{Makhlouf, K.}, \bibinfo{author}{Zhioua, S.}, \bibinfo{author}{Palamidessi, C.}, \bibinfo{year}{2021}.
\newblock \bibinfo{title}{On the applicability of machine learning fairness notions}.
\newblock \bibinfo{journal}{ACM SIGKDD Explorations Newsletter} \bibinfo{volume}{23}, \bibinfo{pages}{14--23}.
\bibitem[{Mandi et~al.(2023)Mandi, Kotary, Berden, Mulamba, Bucarey, Guns and Fioretto}]{mandi2023decision}
\bibinfo{author}{Mandi, J.}, \bibinfo{author}{Kotary, J.}, \bibinfo{author}{Berden, S.}, \bibinfo{author}{Mulamba, M.}, \bibinfo{author}{Bucarey, V.}, \bibinfo{author}{Guns, T.}, \bibinfo{author}{Fioretto, F.}, \bibinfo{year}{2023}.
\newblock \bibinfo{title}{Decision-focused learning: Foundations, state of the art, benchmark and future opportunities}.
\newblock \bibinfo{journal}{arXiv preprint arXiv:2307.13565} .
\bibitem[{Nabi et~al.(2019)Nabi, Malinsky and Shpitser}]{nabi2019learning}
\bibinfo{author}{Nabi, R.}, \bibinfo{author}{Malinsky, D.}, \bibinfo{author}{Shpitser, I.}, \bibinfo{year}{2019}.
\newblock \bibinfo{title}{Learning optimal fair policies}, in: \bibinfo{booktitle}{International Conference on Machine Learning}, \bibinfo{organization}{PMLR}. pp. \bibinfo{pages}{4674--4682}.
\bibitem[{Neal(2020)}]{neal2020introduction}
\bibinfo{author}{Neal, B.}, \bibinfo{year}{2020}.
\newblock \bibinfo{title}{Introduction to causal inference}.
\newblock \bibinfo{journal}{Course Lecture Notes (draft)} \bibinfo{volume}{132}.
\bibitem[{Nie et~al.(2021)Nie, Ye, Liu and Nicolae}]{nie2021vcnet}
\bibinfo{author}{Nie, L.}, \bibinfo{author}{Ye, M.}, \bibinfo{author}{Liu, Q.}, \bibinfo{author}{Nicolae, D.}, \bibinfo{year}{2021}.
\newblock \bibinfo{title}{Vcnet and functional targeted regularization for learning causal effects of continuous treatments}.
\newblock \bibinfo{journal}{arXiv preprint arXiv:2103.07861} .
\bibitem[{Olaya et~al.(2020)Olaya, Coussement and Verbeke}]{olaya2020survey}
\bibinfo{author}{Olaya, D.}, \bibinfo{author}{Coussement, K.}, \bibinfo{author}{Verbeke, W.}, \bibinfo{year}{2020}.
\newblock \bibinfo{title}{A survey and benchmarking study of multitreatment uplift modeling}.
\newblock \bibinfo{journal}{Data Mining and Knowledge Discovery} \bibinfo{volume}{34}, \bibinfo{pages}{273--308}.
\bibitem[{Pearl(2009)}]{pearl2009causal}
\bibinfo{author}{Pearl, J.}, \bibinfo{year}{2009}.
\newblock \bibinfo{title}{{Causal inference in statistics: An overview}}.
\newblock \bibinfo{journal}{Statistics Surveys} \bibinfo{volume}{3}, \bibinfo{pages}{96 -- 146}.
\newblock \DOIprefix\doi{10.1214/09-SS057}.
\bibitem[{Rosenbaum and Rubin(1983)}]{rosenbaum1983central}
\bibinfo{author}{Rosenbaum, P.R.}, \bibinfo{author}{Rubin, D.B.}, \bibinfo{year}{1983}.
\newblock \bibinfo{title}{The central role of the propensity score in observational studies for causal effects}.
\newblock \bibinfo{journal}{Biometrika} \bibinfo{volume}{70}, \bibinfo{pages}{41--55}.
\bibitem[{Rubin(2004)}]{rubin2004direct}
\bibinfo{author}{Rubin, D.B.}, \bibinfo{year}{2004}.
\newblock \bibinfo{title}{Direct and indirect causal effects via potential outcomes}.
\newblock \bibinfo{journal}{Scandinavian Journal of Statistics} \bibinfo{volume}{31}, \bibinfo{pages}{161--170}.
\bibitem[{Rubin(2005)}]{rubin2005causal}
\bibinfo{author}{Rubin, D.B.}, \bibinfo{year}{2005}.
\newblock \bibinfo{title}{Causal inference using potential outcomes: Design, modeling, decisions}.
\newblock \bibinfo{journal}{Journal of the American Statistical Association} \bibinfo{volume}{100}, \bibinfo{pages}{322--331}.
\bibitem[{Scantamburlo et~al.(2024)Scantamburlo, Baumann and Heitz}]{scantamburlo2024prediction}
\bibinfo{author}{Scantamburlo, T.}, \bibinfo{author}{Baumann, J.}, \bibinfo{author}{Heitz, C.}, \bibinfo{year}{2024}.
\newblock \bibinfo{title}{On prediction-modelers and decision-makers: why fairness requires more than a fair prediction model}.
\newblock \bibinfo{journal}{AI \& SOCIETY} , \bibinfo{pages}{1--17}.
\bibitem[{Schr{\"o}der et~al.(2024)Schr{\"o}der, Frauen, Schweisthal, He{\ss}, Melnychuk and Feuerriegel}]{schroder2024conformal}
\bibinfo{author}{Schr{\"o}der, M.}, \bibinfo{author}{Frauen, D.}, \bibinfo{author}{Schweisthal, J.}, \bibinfo{author}{He{\ss}, K.}, \bibinfo{author}{Melnychuk, V.}, \bibinfo{author}{Feuerriegel, S.}, \bibinfo{year}{2024}.
\newblock \bibinfo{title}{Conformal prediction for causal effects of continuous treatments}.
\newblock \bibinfo{journal}{arXiv preprint arXiv:2407.03094} .
\bibitem[{Schwab et~al.(2020)Schwab, Linhardt, Bauer, Buhmann and Karlen}]{schwab2020learning}
\bibinfo{author}{Schwab, P.}, \bibinfo{author}{Linhardt, L.}, \bibinfo{author}{Bauer, S.}, \bibinfo{author}{Buhmann, J.M.}, \bibinfo{author}{Karlen, W.}, \bibinfo{year}{2020}.
\newblock \bibinfo{title}{Learning counterfactual representations for estimating individual dose-response curves}, in: \bibinfo{booktitle}{Proceedings of the AAAI Conference on Artificial Intelligence}, pp. \bibinfo{pages}{5612--5619}.
\bibitem[{Shalit et~al.(2017)Shalit, Johansson and Sontag}]{shalit2017estimating}
\bibinfo{author}{Shalit, U.}, \bibinfo{author}{Johansson, F.D.}, \bibinfo{author}{Sontag, D.}, \bibinfo{year}{2017}.
\newblock \bibinfo{title}{Estimating individual treatment effect: generalization bounds and algorithms}, in: \bibinfo{booktitle}{International conference on machine learning}, \bibinfo{organization}{PMLR}. pp. \bibinfo{pages}{3076--3085}.
\bibitem[{Silva(2016)}]{silva2016observational}
\bibinfo{author}{Silva, R.}, \bibinfo{year}{2016}.
\newblock \bibinfo{title}{Observational-interventional priors for dose-response learning}, in: \bibinfo{editor}{Lee, D.}, \bibinfo{editor}{Sugiyama, M.}, \bibinfo{editor}{Luxburg, U.}, \bibinfo{editor}{Guyon, I.}, \bibinfo{editor}{Garnett, R.} (Eds.), \bibinfo{booktitle}{Advances in Neural Information Processing Systems}, \bibinfo{publisher}{Curran Associates, Inc.}
\newblock \URLprefix \url{https://proceedings.neurips.cc/paper_files/paper/2016/file/aff1621254f7c1be92f64550478c56e6-Paper.pdf}.
\bibitem[{Vanderschueren et~al.(2023a)Vanderschueren, Berrevoets and Verbeke}]{vanderschueren2023noflite}
\bibinfo{author}{Vanderschueren, T.}, \bibinfo{author}{Berrevoets, J.}, \bibinfo{author}{Verbeke, W.}, \bibinfo{year}{2023}a.
\newblock \bibinfo{title}{Noflite: Learning to predict individual treatment effect distributions}.
\newblock \bibinfo{journal}{Transactions on Machine Learning Research} .
\bibitem[{Vanderschueren et~al.(2023b)Vanderschueren, Boute, Verdonck, Baesens and Verbeke}]{vanderschueren2023optimizing}
\bibinfo{author}{Vanderschueren, T.}, \bibinfo{author}{Boute, R.}, \bibinfo{author}{Verdonck, T.}, \bibinfo{author}{Baesens, B.}, \bibinfo{author}{Verbeke, W.}, \bibinfo{year}{2023}b.
\newblock \bibinfo{title}{Optimizing the preventive maintenance frequency with causal machine learning}.
\newblock \bibinfo{journal}{International Journal of Production Economics} \bibinfo{volume}{258}, \bibinfo{pages}{108798}.
\bibitem[{Vanderschueren et~al.(2024)Vanderschueren, Verbeke, Moraes and Proen{\c{c}}a}]{vanderschueren2024metalearners}
\bibinfo{author}{Vanderschueren, T.}, \bibinfo{author}{Verbeke, W.}, \bibinfo{author}{Moraes, F.}, \bibinfo{author}{Proen{\c{c}}a, H.M.}, \bibinfo{year}{2024}.
\newblock \bibinfo{title}{Metalearners for ranking treatment effects}.
\newblock \bibinfo{journal}{arXiv preprint arXiv:2405.02183} .
\bibitem[{Vasquez et~al.(2022)Vasquez, De~Weerdt and vanden Broucke}]{vasquez2022hidden}
\bibinfo{author}{Vasquez, C.O.}, \bibinfo{author}{De~Weerdt, J.}, \bibinfo{author}{vanden Broucke, S.}, \bibinfo{year}{2022}.
\newblock \bibinfo{title}{The hidden cost of fraud: An instance-dependent cost-sensitive approach for positive and unlabeled learning}, in: \bibinfo{booktitle}{Fourth International Workshop on Learning with Imbalanced Domains: Theory and Applications}, \bibinfo{organization}{PMLR}. pp. \bibinfo{pages}{53--67}.
\bibitem[{Verbeke et~al.(2012)Verbeke, Dejaeger, Martens, Hur and Baesens}]{verbeke2012new}
\bibinfo{author}{Verbeke, W.}, \bibinfo{author}{Dejaeger, K.}, \bibinfo{author}{Martens, D.}, \bibinfo{author}{Hur, J.}, \bibinfo{author}{Baesens, B.}, \bibinfo{year}{2012}.
\newblock \bibinfo{title}{New insights into churn prediction in the telecommunication sector: A profit driven data mining approach}.
\newblock \bibinfo{journal}{European Journal of Operational Research} \bibinfo{volume}{218}, \bibinfo{pages}{211--229}.
\bibitem[{Verbeke et~al.(2023)Verbeke, Olaya, Guerry and Van~Belle}]{verbeke2023or}
\bibinfo{author}{Verbeke, W.}, \bibinfo{author}{Olaya, D.}, \bibinfo{author}{Guerry, M.A.}, \bibinfo{author}{Van~Belle, J.}, \bibinfo{year}{2023}.
\newblock \bibinfo{title}{To do or not to do? cost-sensitive causal classification with individual treatment effect estimates}.
\newblock \bibinfo{journal}{European Journal of Operational Research} \bibinfo{volume}{305}, \bibinfo{pages}{838--852}.
\bibitem[{Verbraken et~al.(2012)Verbraken, Verbeke and Baesens}]{verbraken2012novel}
\bibinfo{author}{Verbraken, T.}, \bibinfo{author}{Verbeke, W.}, \bibinfo{author}{Baesens, B.}, \bibinfo{year}{2012}.
\newblock \bibinfo{title}{A novel profit maximizing metric for measuring classification performance of customer churn prediction models}.
\newblock \bibinfo{journal}{IEEE transactions on knowledge and data engineering} \bibinfo{volume}{25}, \bibinfo{pages}{961--973}.
\bibitem[{Verstraete et~al.(2023)Verstraete, Gyselinck, Huts, Das, Topalovic, De~Vos and Janssens}]{verstraete2023estimating}
\bibinfo{author}{Verstraete, K.}, \bibinfo{author}{Gyselinck, I.}, \bibinfo{author}{Huts, H.}, \bibinfo{author}{Das, N.}, \bibinfo{author}{Topalovic, M.}, \bibinfo{author}{De~Vos, M.}, \bibinfo{author}{Janssens, W.}, \bibinfo{year}{2023}.
\newblock \bibinfo{title}{Estimating individual treatment effects on copd exacerbations by causal machine learning on randomised controlled trials}.
\newblock \bibinfo{journal}{Thorax} \bibinfo{volume}{78}, \bibinfo{pages}{983--989}.
\newblock \URLprefix \url{https://thorax.bmj.com/content/78/10/983}, \DOIprefix\doi{10.1136/thorax-2022-219382}, \href{http://arxiv.org/abs/https://thorax.bmj.com/content/78/10/983.full.pdf}{{\tt arXiv:https://thorax.bmj.com/content/78/10/983.full.pdf}}.
\bibitem[{Zhan et~al.(2024)Zhan, Liu, Li and Wu}]{zhan2024weighted}
\bibinfo{author}{Zhan, B.}, \bibinfo{author}{Liu, C.}, \bibinfo{author}{Li, Y.}, \bibinfo{author}{Wu, C.}, \bibinfo{year}{2024}.
\newblock \bibinfo{title}{Weighted doubly robust learning: An uplift modeling technique for estimating mixed treatments' effect}.
\newblock \bibinfo{journal}{Decision Support Systems} \bibinfo{volume}{176}, \bibinfo{pages}{114060}.
\bibitem[{Zhang et~al.(2021)Zhang, Li and Liu}]{zhang2021unified}
\bibinfo{author}{Zhang, W.}, \bibinfo{author}{Li, J.}, \bibinfo{author}{Liu, L.}, \bibinfo{year}{2021}.
\newblock \bibinfo{title}{A unified survey of treatment effect heterogeneity modelling and uplift modelling}.
\newblock \bibinfo{journal}{ACM Computing Surveys (CSUR)} \bibinfo{volume}{54}, \bibinfo{pages}{1--36}.
\bibitem[{Zhang et~al.(2022)Zhang, Zhang, Lipton, Li and Xing}]{zhang2022exploring}
\bibinfo{author}{Zhang, Y.F.}, \bibinfo{author}{Zhang, H.}, \bibinfo{author}{Lipton, Z.C.}, \bibinfo{author}{Li, L.E.}, \bibinfo{author}{Xing, E.P.}, \bibinfo{year}{2022}.
\newblock \bibinfo{title}{Exploring transformer backbones for heterogeneous treatment effect estimation}.
\newblock \bibinfo{journal}{arXiv preprint arXiv:2202.01336} .
\bibitem[{Zhao et~al.(2024)Zhao, Bai, Xiong, Cao, Ma, Jiang, Wu and Kuang}]{zhao2024learning}
\bibinfo{author}{Zhao, Z.}, \bibinfo{author}{Bai, Y.}, \bibinfo{author}{Xiong, R.}, \bibinfo{author}{Cao, Q.}, \bibinfo{author}{Ma, C.}, \bibinfo{author}{Jiang, N.}, \bibinfo{author}{Wu, F.}, \bibinfo{author}{Kuang, K.}, \bibinfo{year}{2024}.
\newblock \bibinfo{title}{Learning individual treatment effects under heterogeneous interference in networks}.
\newblock \bibinfo{journal}{ACM Transactions on Knowledge Discovery from Data} \bibinfo{volume}{18}, \bibinfo{pages}{1--21}.
\bibitem[{Zhou et~al.(2023)Zhou, Li, Jiang, Zheng and Wang}]{zhou2023direct}
\bibinfo{author}{Zhou, H.}, \bibinfo{author}{Li, S.}, \bibinfo{author}{Jiang, G.}, \bibinfo{author}{Zheng, J.}, \bibinfo{author}{Wang, D.}, \bibinfo{year}{2023}.
\newblock \bibinfo{title}{Direct heterogeneous causal learning for resource allocation problems in marketing}, in: \bibinfo{booktitle}{Proceedings of the AAAI Conference on Artificial Intelligence}, pp. \bibinfo{pages}{5446--5454}.

\end{thebibliography}


\end{document}